\documentclass[final,5p,times,twocolumn,authoryear]{elsarticle}

\usepackage{amssymb}
\usepackage{amsmath}
\usepackage{amsthm}
\usepackage{hyperref}
\DeclareMathOperator{\Var}{Var}
\newcommand{\wbar}{\overline}
\newcommand{\R}{\mathbb{R}}

\usepackage{lineno}
\journal{ISPRS Journal of Photogrammetry and Remote Sensing}

\begin{document}

\begin{frontmatter}

%% Title, authors and addresses

%% use the tnoteref command within \title for footnotes;
%% use the tnotetext command for theassociated footnote;
%% use the fnref command within \author or \affiliation for footnotes;
%% use the fntext command for theassociated footnote;
%% use the corref command within \author for corresponding author footnotes;
%% use the cortext command for theassociated footnote;
%% use the ead command for the email address,
%% and the form \ead[url] for the home page:
%% \title{Title\tnoteref{label1}}
%% \tnotetext[label1]{}
%% \author{Name\corref{cor1}\fnref{label2}}
%% \ead{email address}
%% \ead[url]{home page}
%% \fntext[label2]{}
%% \cortext[cor1]{}
%% \affiliation{organization={},
%%            addressline={}, 
%%            city={},
%%            postcode={}, 
%%            state={},
%%            country={}}
%% \fntext[label3]{}

\title{PoseIDON: 6DoF Pose Estimation with Foundation Model Features for Marine Sediment Burial Mapping} %% Article title

%% use optional labels to link authors explicitly to addresses:
%% \author[label1,label2]{}
%% \affiliation[label1]{organization={},
%%             addressline={},
%%             city={},
%%             postcode={},
%%             state={},
%%             country={}}
%%
%% \affiliation[label2]{organization={},
%%             addressline={},
%%             city={},
%%             postcode={},
%%             state={},
%%             country={}}

\author[mpl,ece]{Jerry Yan\corref{cor1}} %% Author name
\ead{jeyan@ucsd.edu}
\cortext[cor1]{Corresponding author}
\author[ece]{Chinmay Talegaonkar}
\author[ece]{Nicholas Antipa}
\author[mpl]{Eric Terrill}
\author[mpl]{Sophia Merrifield}

%% Author affiliation
\affiliation[mpl]{organization={Marine Physical Laboratory, Scripps Institution of Oceanography},%Department and Organization
            addressline={UCSD}, 
            city={La Jolla},
            postcode={92037}, 
            state={CA},
            country={USA}}
\affiliation[ece]{organization={Department of Electrical and Computer Engineering},%Department and Organization
            addressline={UCSD}, 
            city={La Jolla},
            postcode={92037}, 
            state={CA},
            country={USA}}

%% Abstract
\begin{abstract}
%% Text of abstract
The burial state of anthropogenic objects on the seafloor provides insight into localized sedimentation dynamics and is also critical for assessing ecological risks, potential pollutant transport, and the viability of recovery or mitigation strategies for hazardous materials such as munitions. Accurate burial depth estimation from remote imagery remains difficult due to partial occlusion, poor visibility, and object degradation. This work introduces a computer vision pipeline, called PoseIDON, which combines deep foundation model features with multiview photogrammetry to estimate six degrees of freedom object pose and the orientation of the surrounding seafloor from ROV video. Burial depth is inferred by aligning CAD models of the objects with observed imagery and fitting a local planar approximation of the seafloor. The method is validated using footage of 54 objects, including barrels and munitions, recorded at a historic ocean dumpsite in the San Pedro Basin. The model achieves a mean burial depth error of approximately 10 centimeters and resolves spatial burial patterns that reflect underlying sediment transport processes. This approach enables scalable, non-invasive mapping of seafloor burial and supports environmental assessment at contaminated sites.
\end{abstract}

%%Graphical abstract
% \begin{graphicalabstract}
%\includegraphics{grabs}
% \end{graphicalabstract}

%%Research highlights
% \begin{highlights}
% \item Research highlight 1
% \item Research highlight 2
% \end{highlights}

%% Keywords
\begin{keyword}
%% keywords here, in the form: keyword \sep keyword

%% PACS codes here, in the form: \PACS code \sep code

%% MSC codes here, in the form: \MSC code \sep code
%% or \MSC[2008] code \sep code (2000 is the default)
pose estimation \sep marine debris \sep seafloor mapping \sep robotics monitoring \sep computer vision \sep unexploded ordnance
\end{keyword}

\end{frontmatter}

%% Add \usepackage{lineno} before \begin{document} and uncomment 
%% following line to enable line numbers
%% \linenumbers

%% main text
%%

\section{Introduction}
The mapping and imaging of man-made objects on the seafloor are important tools for the inspection of marine debris fields, seabed infrastructure and cultural preservation. While seabed imaging techniques have evolved in the past few decades, challenges remain in the estimation of the burial state of objects, due in part to object occlusion from sediment. A new approach is developed that leverages deep learning techniques and \textit{a-priori} knowledge of the object to make informed estimates of the burial state of man-made objects. This study focuses on objects imaged during a wide area survey of a historic dumpsite in the Southern California San Pedro Basin \citep{merrifield_wide-area_2023} via a remotely operated vehicle (ROV). These dumpsites are oceanic regions where hazardous industrial byproducts were dumped and pose risks to the marine food web and human health \citep{mackintosh_newly_2016}. In particular, this survey is of a known dumpsite between 1947 and 1961 for the chemical dichlorodiphenyltrichloroethane (DDT), which has been detected extensively in the basin's sediment as a result of bulk dumping \citep{schmidt_disentangling_2024}. The dumpsite also has a large number of distributed, barrel-sized ($\sim$1m) objects that were determined to be both containerized waste and discarded munitions \citep{merrifield_wide-area_2023}.

To characterize these objects, the position, orientation, and burial depth are measured by performing 6 degrees of freedom (6DoF) pose estimation. This information provides important initial conditions for predicting object mobility \citep{chu_prediction_2022}, can help make inferences on object composition (burial depth can imply object weight, e.g. empty containers will be shallower), and can make estimates on marine sedimentation rates. For measuring marine sedimentation rates in particular, standard methods require in situ sample collection of sediment and months to years of laboratory testing \citep{restreppo_oceanic_2020}. Using computer vision and \textit{a-priori} knowledge of a buried object, 6DoF pose can be estimated using only camera data, enabling a technique minimally invasive to the environment to describe these objects.

Most seabed vision research focuses on object detection, classification \citep{chen_new_2021, katija_fathomnet_2022}, and ROV navigation through visual pose estimation \citep{nielsen_evaluation_2019, kim_real-time_2013, zhang_visual_2022}, while rigid object pose estimation has been limited \citep{yan_pose_2024}. Submerged objects visually deteriorate over time, causing significant variation even within the same category. Thus, reference CAD models must be generic and untextured, making point matching between the reference model and the real object challenging. This study shows that recent vision foundation models like DINOv2 \citep{oquab_dinov2_2024}, which learn versatile visual features, can match synthetic CAD renderings to their real, degraded underwater counterparts without retraining.

This can enable robust pose estimation from a single image in a variety of environments. However, underwater conditions still make this problem challenging for DINOv2, as there were significantly more errors in its predictions. To filter out these outliers, classical multiview techniques \citep{schonberger_structure--motion_2016, schonberger_pixelwise_2016} can be used to find geometric inconsistencies. Finally, since the CAD model is scaled, the 3D reconstruction can be properly scaled without control points in the scene.

This work will introduce PoseIDON (\textbf{Pose} \textbf{I}dentification for \textbf{D}epth of \textbf{O}bjects via foundation model \textbf{N}etworks), a 6DoF pose estimation model for buried objects that overcomes the data quality and quantity limitations of ROV footage. Foundation features are leveraged without the need for retraining, in combination with classical multiview methods to enforce 3D geometric consistency, localize the seafloor relative to the buried object, and calculate burial depth using \textit{a-priori} object dimensions. This allows the calculation of burial depth of any object from arbitrary ROV footage without training, given a CAD model of it exists. The potential of this approach is demonstrated quantitatively, using labeled ROV camera footage of buried objects found at a coastal historic dump site approximately 350 $\textrm{km}^2$ in area. Source code and example imagery are available at \url{https://github.com/jerukan/PoseIDON}.

\section{Related Work}
This paper adapts research from classical pose estimation methods and recent deep learning models in computer vision.

\subsection{6DoF Pose Estimation}
Current work in pose estimation is generally split into the following categories: RGB-D methods, which comprise color (RGB) and depth (D) of each scene point, and RGB-only methods, where only color is available. RGB-D methods usually operate on point clouds directly and are generally more accurate than RGB methods \citep{liu_deep_2024}. Early classical methods used hand-crafted features in image space \citep{collet_moped_2011, lowe_object_1999} or point cloud space \citep{besl_method_1992, sahillioglu_scale-adaptive_2021}, but modern approaches use deep learning and are more robust to different environments. The data in this study comprise RGB footage of submerged objects, captured as the ROV operator navigates around them. This provides multiple view directions of the objects, which is used to infer structure via 3D reconstruction. However, the footage was arbitrarily recorded and lacks many qualities preferred for 3D reconstruction: almost all footage did not have a 360 degree view coverage, the lighting is non-uniform, visibility is limited from haze, and suspended particles like marine snow are prevalent in the footage. Metadata for camera intrinsics were also not recorded with the footage, meaning additional camera parameters need to be estimated. This degrades the quality of 3D reconstructions, and along with inherent scale ambiguity, these factors make RGB-D methods less reliable. Preceding work \citep{yan_pose_2024} attempted pose estimation from unscaled point clouds via deep learning, but it was limited to only predicting the pose of a single object type (barrels), and could not recover absolute scene scale. Hence, a method is needed that needs only RGB fly-by data to estimate 6DoF pose. 

Deep learning RGB methods themselves are further separated into three categories \citep{liu_deep_2024}. The first type are instance-level methods \citep{guo_knowledge_2023, xu_rnnpose_2024}, which can only predict the poses of objects they were trained on. This is not scalable as it requires retraining for new objects, and this study lacks the data quantity needed. The second type are category-level methods \citep{fan_object_2022, wei_rgb-based_2024}, where networks are trained to estimate the poses of similar objects in one category, potentially resolving the deterioration differences between objects. However, these methods rely heavily on monocular depth estimates and also require training for each category. High quality ground truth 3D data does not exist for the current study at scale, making category-level methods infeasible. The final type are unseen object methods \citep{nguyen_gigapose_2024, ausserlechner_zs6d_2024, ornek_foundpose_2024}, which predict the pose of any object given a reference CAD model. With the introduction of vision transformers \citep{dosovitskiy_image_2021}, foundation models have appeared, allowing some unseen object methods to work without task-specific training \citep{ausserlechner_zs6d_2024, ornek_foundpose_2024}. Foundation models are large models that are trained on a vast quantity of data to produce high-dimensional features that are suitable for many computer vision tasks. These features were found to be robust across different imaging domains, and often performed better than task-specific models \citep{bommasani_opportunities_2022, caron_emerging_2021}. These deep learning pose estimation models have some robustness to object occlusion, but have not been tested extensively in extreme conditions such as underwater imaging.

\subsection{Underwater 3D Reconstruction}
The industry standard method for 3D reconstruction is classical photogrammetry, which consists of two important steps. It first jointly estimates a sparse point cloud and camera positions in a process called structure-from-motion (SfM) \citep{schonberger_structure--motion_2016}, which is then densified into a dense point cloud and turned into a texture mesh via multiview stereo (MVS) \citep{schonberger_pixelwise_2016}. This entire pipeline is known as multiview geometry (MVG). Photogrammetry is used extensively for terrestrial and aerial surveys \citep{agarwal_building_2011, svennevig_oblique_2015,schonberger_structure--motion_2016}, and has been adapted for underwater surveys \citep{arnaubec_underwater_2023, sedlazeck_3d_2009, pizarro_large_2004, drap_underwater_2013}.

Recent advances in deep neural networks have led to many new 3D reconstruction methods designed to be more accurate and robust than classical photogrammetry. Many methods replace individual parts of the MVG pipeline with neural networks \citep{lindenberger_pixel-perfect_2021, vijayanarasimhan_sfm-net_2017}, while the newest methods completely replace the MVG pipeline and predict 3D structure directly from RGB images \citep{yang_fast3r_2025, wang_vggt_2025}. While these data-driven models are generally more robust than classical methods, the overwhelming majority of 3D data are terrestrial, and underwater 3D datasets are not available at a large scale. This makes retraining the models specifically for underwater data unviable, and their performances suffer as a result.

\section{Materials and Methods}
\subsection{Problem Formulation}
\begin{figure}
    \centering
    \includegraphics[width=.8\linewidth]{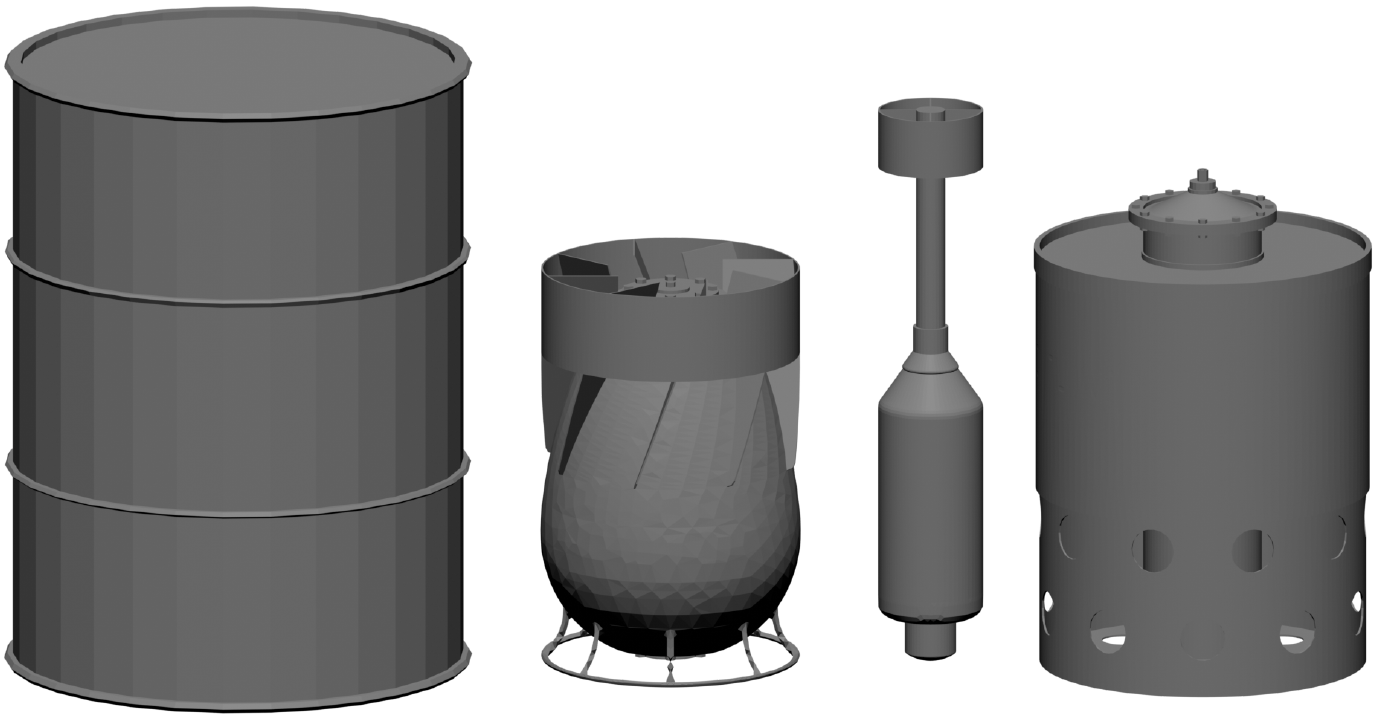}
    \caption{To-scale untextured renderings of the CAD models fed into FoundPose. From left to right: drum barrel, depth charge, mousetrap, and smoke float.}
    \label{fig:cad-models}
\end{figure}

The problem is defined as the estimation of the 6DoF pose of a buried object and the seafloor from multiple deep-sea RGB images in order to calculate its burial depth, where the seafloor is modeled as a plane. A to-scale CAD model of the object is available as a reference, but is untextured, since real objects are in various states of degradation. The objects being tracked are listed in \autoref{tab:object-table}, and their corresponding CAD models are shown in \autoref{fig:cad-models}. Each observed object has their latitude and longitude recorded, allowing the mapping of burial depths.
\begin{table*}
    \centering
    \caption{The list of target objects analyzed with their corresponding sizes and wet weights.}
    \begin{tabular}{|c|c|c|c|}
        \hline
        Object & Diameter (cm) & Height (cm) & Wet Weight (kg) \\
        \hline\hline
        110 gal. drum barrel & 76.2 & 106.68 & 28.5 \\
        Mark 9 depth charge mod 1 & 44.81 & 70.17 & 79.3 \\
        Mousetrap rocket & 18.3 & 99.06 & 178.5 \\
        HC smoke float mark 1 & 57.15 & 77.98 & 15.7 \\
        \hline
    \end{tabular}
    \label{tab:object-table}
\end{table*}

\subsection{Data Collection}
\label{sec:mtd-datacollect}
The data used in this experiment were recorded over the course of several ROV dives at known dump sites in the San Pedro Basin in 2023 \citep{merrifield_wide-area_2023}. The ROV used was the CURV-21, which was equipped with an Imenco OE14-504 Wide Angle Camera and two DeepSea lights. Most footage was collected at a depth around 1000 meters, recorded at an altitude 1-4 meters above the seafloor, and illuminated with a single forward-facing light source. The recording was done by a wide angle camera, but due to difficulties, its intrinsics are unknown. The camera zoom levels changed during dives, and its metadata was not recorded.

Video clips of buried objects were chosen if they were recorded close enough to the object for COLMAP to reconstruct, which meant around 5 meters away or closer. Each video clip of an object lasts around 30 seconds to 2 minutes, and frames of the video were extracted every 1-2 seconds, giving us around 30-60 images per object.

Although 1486 man-made objects were observed over the course of multiple dives, 316 of which were barrels, only a highly limited selection of 54 objects were utilized due to labeling difficulties.

\subsection{Preliminaries}
This section provides the necessary preliminary information to understand the methodology.
\subsubsection{Representing Object Poses and Cameras}
For object position and orientation, standard notation is used to represent them in a single transformation: $\mathbf{T}\in \textrm{SE}(3) \subset \R^{4\times4}$, which is parametrized by a rotation matrix $\mathbf{R}\in \textrm{SO}(3) \subset \R^{3\times3}$ and translation vector $\mathbf{t}\in\R^3$. When a transformation is performed, the rotation is applied first, then the translation, as indicated by the following notation:
\begin{equation}
    \mathbf{T} = [\mathbf{R}|\mathbf{t}] = \begin{bmatrix}
        \mathbf{R} & \mathbf{t} \\
        \mathbf{0}^\top & 1
    \end{bmatrix}.
    \label{eq:transform}
\end{equation}
A transformation of $a$ relative to $b$ is denoted by $_{b}\mathbf{T}_a$.

The cameras follow the standard pinhole model with square pixels and zero skew. A point $[x,y,z]^\top$ relative to the camera frame is projected onto pixel coordinates $[u,v]^\top$ in image space via the following:
\begin{equation}
\begin{split}
    \begin{bmatrix}
        u\\v\\1
    \end{bmatrix}= \frac{1}{z}\mathbf{K}\begin{bmatrix}
        x\\y\\z
    \end{bmatrix} \qquad    
    \mathbf{K} = \begin{bmatrix}
        f & 0 & c_x \\
        0 & f & c_y \\
        0 & 0 & 1
    \end{bmatrix}
    \end{split},
    \label{eq:intrinsics}
\end{equation}
where $\mathbf{K}$ is the camera intrinsic matrix, $f$ is the camera's focal length, and $(c_x,c_y)$ is the camera's principal point.

\subsubsection{Robust Feature Detection and Matching}
\label{sec:dinov2-robust}
\begin{figure}[h!]
    \centering
    \includegraphics[width=.8\linewidth]{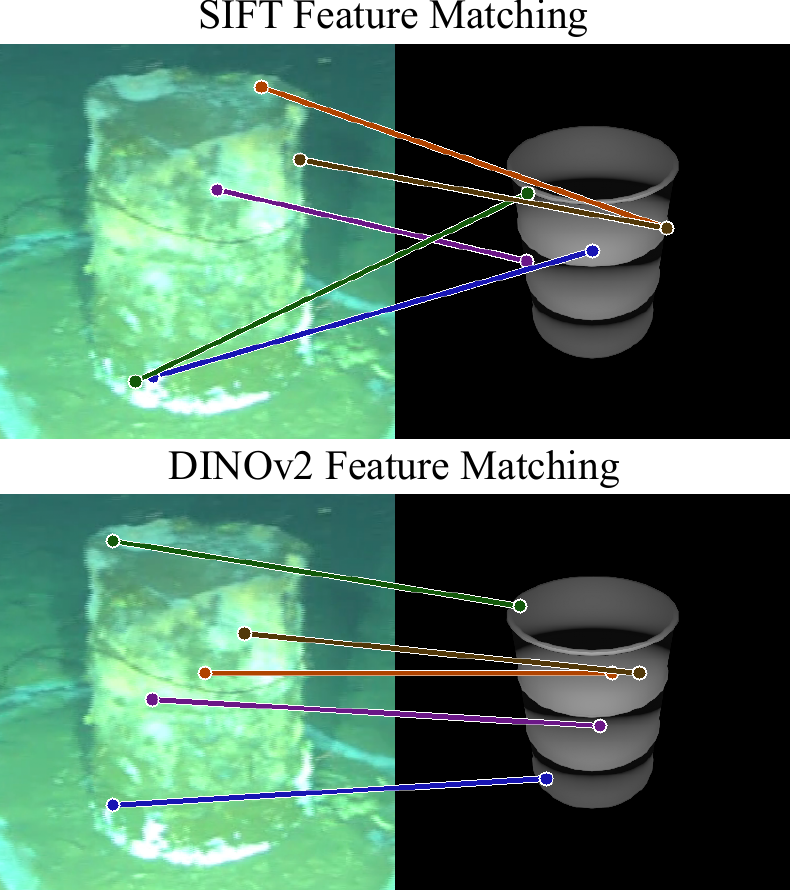}
    \caption{A comparison of feature matching between hand-crafted SIFT features \citep{lowe_distinctive_2004} (top), and DINOv2 foundation features \citep{oquab_dinov2_2024} (bottom). Features are matched between an upright, mostly unburied barrel (left) and its most similar CAD model render (right). SIFT's inability to match features between image domains is apparent, while DINOv2 considers the global context of each feature to produce fairly decent matches.}
    \label{fig:siftvsdino}
\end{figure}

Image feature detection and matching are fundamental computer vision tasks that involve identifying key image regions and linking similar areas between different viewpoints of the same subject. These matches are defined as feature correspondences. Classical methods involve hand-crafted features, such as SIFT \citep{lowe_distinctive_2004}, which look for distinctive features such as corners in a local context. When creating correspondences, these matches are robust to noise, different viewing angles, and scales, but often fail with textureless scenes and differing environmental conditions between images (\autoref{fig:siftvsdino}). When attempting to find correspondences between inconsistent environments, a more robust method is needed.

Recent deep learning methods can address these shortcomings in robustness. Specifically, DINOv2  \citep{oquab_dinov2_2024}, a foundation model built on the vision transformer (ViT) \citep{dosovitskiy_image_2021}, is used in this work. It has been shown to encode both local spatial information about object parts, and global semantic information of object categories \citep{amir_deep_2022}. In particular, a frozen intermediate layer of DINOv2 produces a mix of positional and semantic information for different image regions that can produce geometrically consistent correspondences \citep{ornek_foundpose_2024}. This important property allows the establishment of synthetic-to-real correspondences (\autoref{fig:siftvsdino}). In the scenes themselves, objects of the same type across different videos have both texture and geometric inconsistencies, so DINOv2 allows features to be matched, regardless of the object's state of decay underwater, without any need for training.
\subsubsection{Enforcing Geometric Consistency}
\begin{figure}
    \centering
    \includegraphics[width=.6\linewidth]{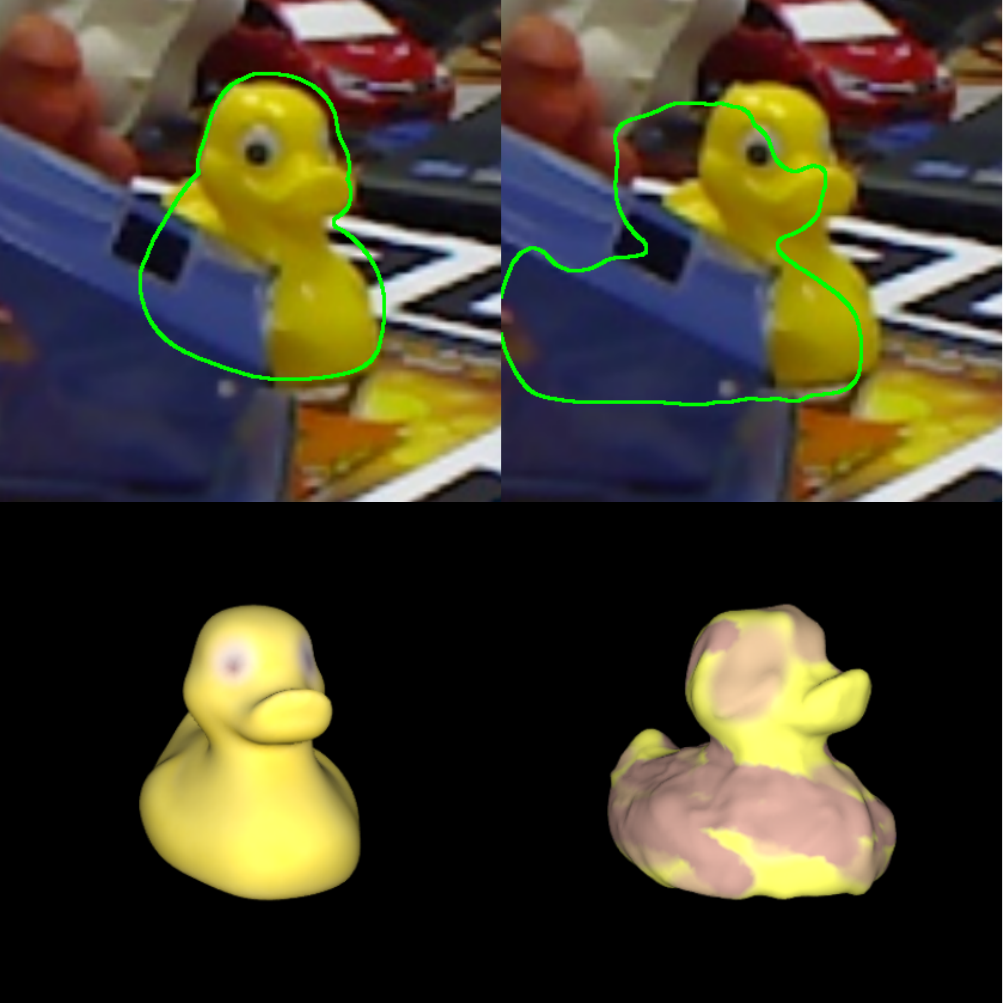}
    \caption{A toy example of how texture inconsistencies can affect the performance of FoundPose \citep{ornek_foundpose_2024}. The highest scoring template for the default duck (bottom left) was correctly selected, but the highest scoring template for the rusty duck (bottom right) is incorrect. This leads to an accurate pose for the default duck (top left) but an incorrect pose for the rusty duck (top right). This means lower-scoring templates must be considered, and a voting method must be used to exclude outlier poses.}
    \label{fig:rustyduck}
\end{figure}

\begin{figure}[h!]
    \centering
    \includegraphics[width=.85\linewidth]{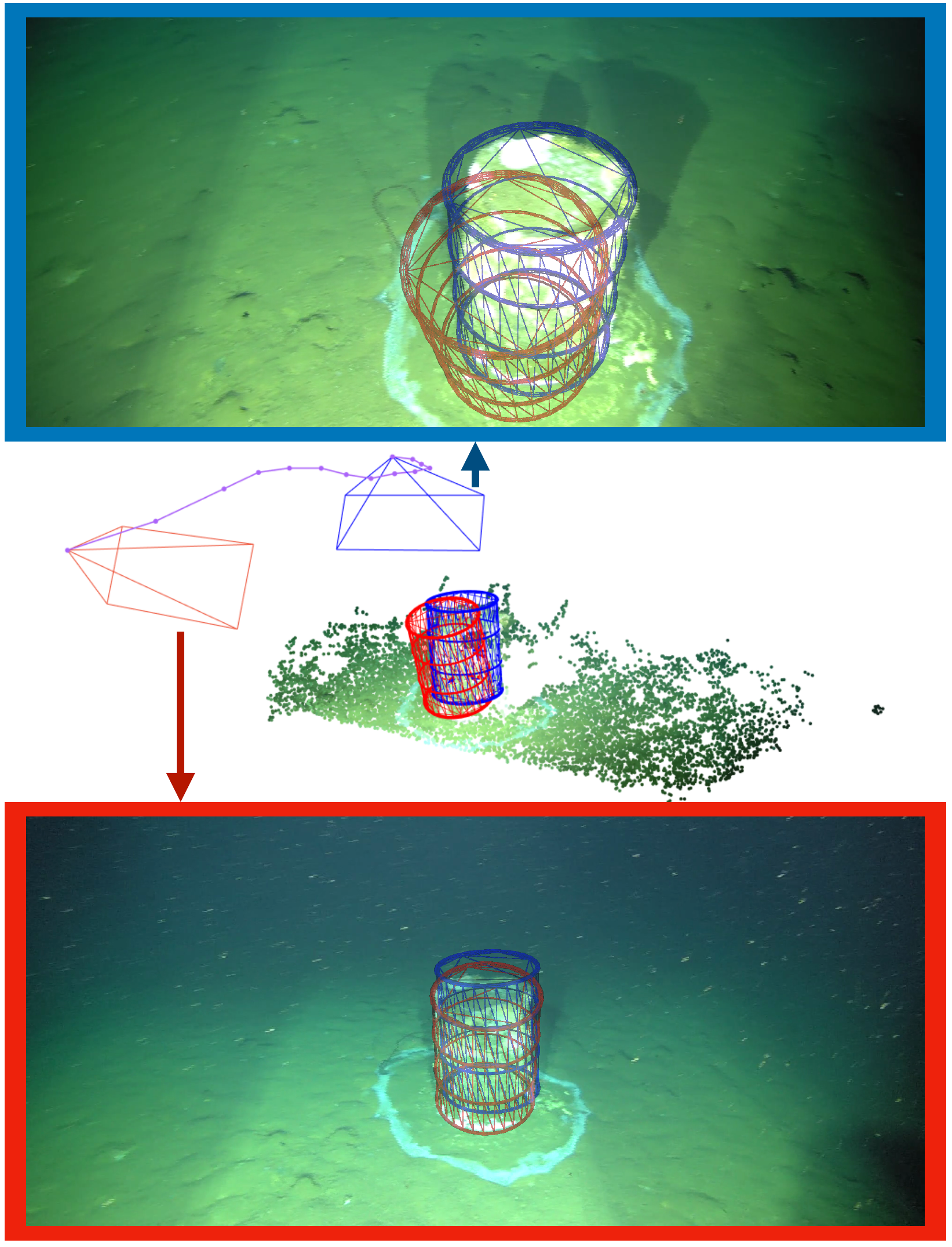}
    \caption{Video frames are shown from the blue camera (top) and red camera (bottom), and the camera's movement trajectory between them (middle purple). Relative barrel poses from each camera in view of the other are shown as wireframes. The barrel pose predicted by the red camera is not visually incorrect in its own view (bottom red wireframe), but when viewed from the blue camera, it is clearly incorrect (top red wireframe), and vice versa with the blue camera.}
    \label{fig:multiview-important}
\end{figure}

Current state-of-the-art RGB pose estimators like FoundPose \citep{ornek_foundpose_2024} generally perform well for monocular pose estimation, but texture and geometry inconsistencies between the CAD model and real object can greatly affect performance (\autoref{fig:rustyduck}). Additional object images from multiview data help enforce geometric consistency as seen in \autoref{fig:multiview-important}. By viewing the monocular predictions of one camera from the perspective of another camera, it becomes easier to detect outlier pose predictions.
\subsubsection{Outlier Detection with RANSAC}
Due to the large number of extraneous features in underwater images such as marine snow, biological growth, and distinct shadows from ROV lighting, outliers often appear in most of the processing steps. One of the most popular methods for detecting outliers when fitting models is random sample consensus (RANSAC) \citep{fischler_random_1981}. It works by repeatedly taking samples of the data, fitting the model to the sample, and including all points within a certain threshold of error as inliers. The model fitted with the highest number of inliers is chosen as the best model. RANSAC is used in several steps of the model.

\subsection{Overview}
\label{sec:mtd-overview}
\begin{figure*}
    \centering
    \includegraphics[width=.99\linewidth]{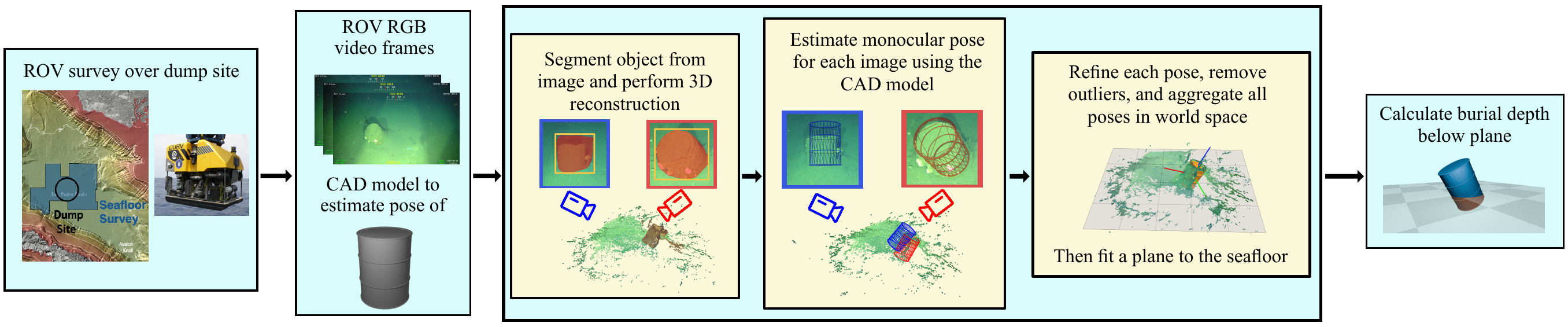}
    \caption{An overview of PoseIDON's pipeline. Given a series of RGB underwater images, inpainting is performed on the text overlay in all images. Grounding DINO \citep{liu_grounding_2023} is used to find the bounding box of the object, and SAM \citep{ravi_sam_2024} to extract the object mask. The images, masks, and reference CAD model are then fed into FoundPose \citep{ornek_foundpose_2024} to predict monocular poses relative to each camera. Unscaled camera world positions are then recovered and a 3D point cloud of the scene is determined via photogrammetry \citep{schonberger_structure--motion_2016, schonberger_pixelwise_2016}, with the object and seafloor points segmented using the masks. The scale ambiguity of the camera positions and point cloud is resolved using the coarse FoundPose estimates, followed by ICP and pose averaging to refine the object pose. A plane is fit to the seafloor point cloud to then estimate the object burial depth.}
    \label{fig:pipeline}
\end{figure*}

Images extracted from ROV footage of a buried object are inpainted in sections of the images containing text overlays and ROV hardware. 3D multiview reconstruction is performed via photogrammetry to recover camera positions and a point cloud of the scene, which has an ambiguous scale. A mask of the object is extracted from each image using Grounding DINO \citep{liu_grounding_2023} and Segment Anything Model (SAM) \citep{ravi_sam_2024}, and used to segment the object from the point cloud. The images, masks, and reference CAD model are then fed into FoundPose \citep{ornek_foundpose_2024} to produce a coarse monocular estimation of pose relative to each camera. Using the relative coarse estimates and scale of the CAD model, the camera positions and point cloud are scaled to real-world units. The coarse poses are then refined via iterative closest point (ICP) \citep{besl_method_1992} with the object point cloud, and the best poses are selected via RANSAC \citep{fischler_random_1981} and averaged. A plane is then fit to the points in the point cloud corresponding to the seafloor, from which the object beneath the plane is measured, to then estimate its burial depth. The pipeline of this method is shown in \autoref{fig:pipeline}.

\subsection{Preprocessing}
\label{sec:mtd-preprocess}
\begin{figure}
    \centering
    \includegraphics[width=.8\linewidth]{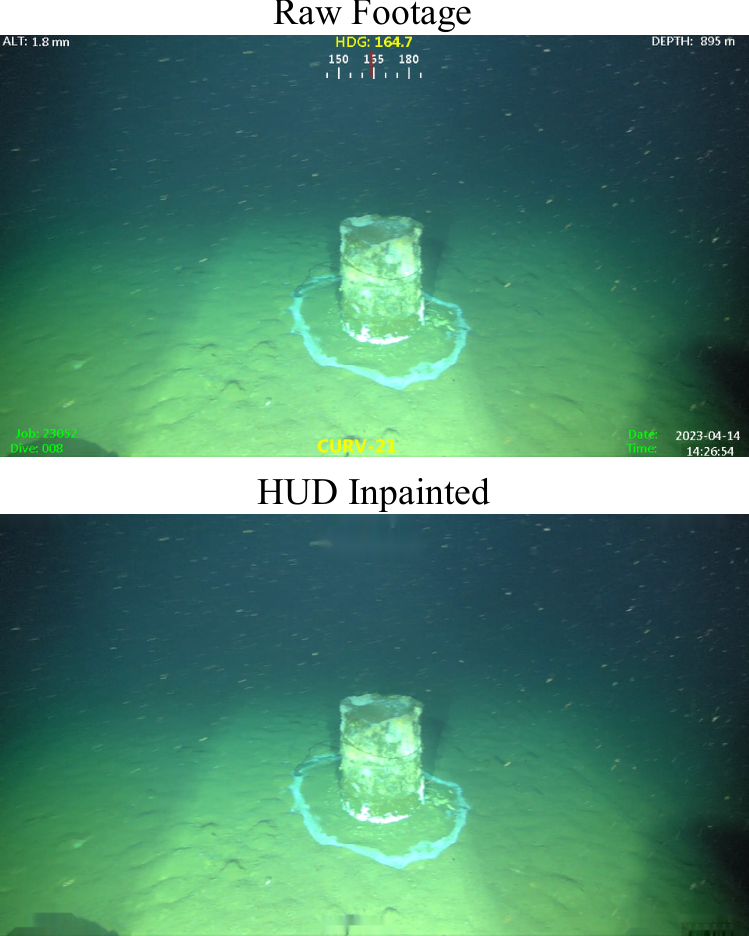}
    \caption{An example video frame, with the original image (top) and the image with the text overlay inpainted (bottom).}
    \label{fig:preprocessing-inpaint}
\end{figure}

ROV footage often has burnt-in text overlays or ROV hardware always visible in frame, which makes 3D reconstruction impossible in some cases since the text or hardware will be detected as features despite not being part of the real scene. This problem can either be resolved with image inpainting or cropping if the inpainting region is too large. Since the data from this study has relatively small text overlays, inpainting was used via the Fast Marching Method \citep{telea_image_2004}, as shown in \autoref{fig:preprocessing-inpaint}. Even if the inpainting is inaccurate, it minimizes the text's interference with feature detection. This provides $N$ preprocessed RGB images $\{I_i\}_{i=1}^N$.

\subsection{Segmentation}
\label{sec:mtd-segment}
\begin{figure}
    \centering
    \includegraphics[width=.8\linewidth]{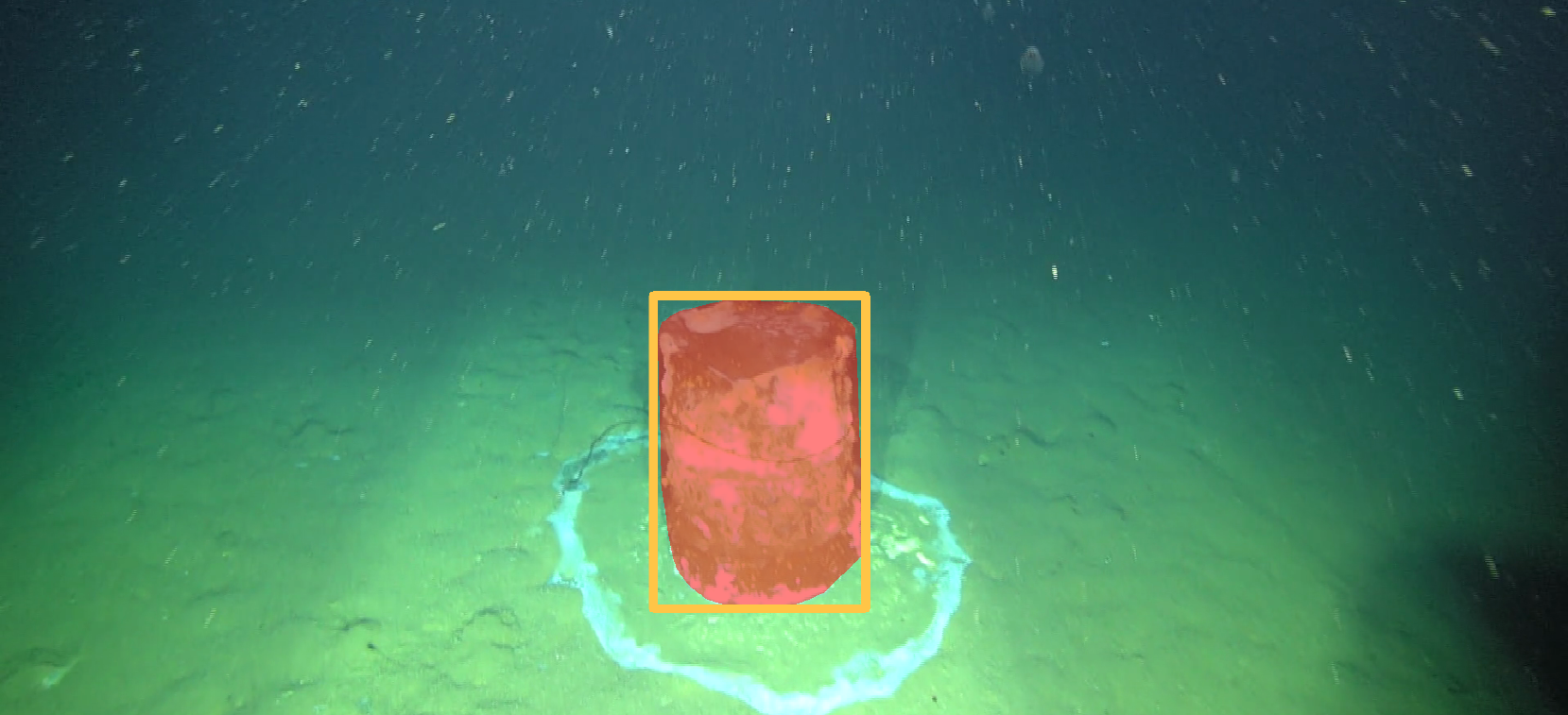}
    \caption{An example video frame with the barrel bounding box (orange) found by Grounding DINO \citep{liu_grounding_2023}, and the mask (red) found by SAM \citep{ravi_sam_2024} after applying a convex hull.}
    \label{fig:segmentation}
\end{figure}

The object is isolated from the background via image segmentation. Since training data at scale is not available in this study, foundation models are employed for both object detection and segmentation. Images are provided to transformer-based object detector Grounding DINO \citep{liu_grounding_2023} with the text prompt ``seabed underwater junk" to extract the bounding box around the target object. Due to unrelated objects like jellyfish sometimes appearing in the video, multiple objects can be mistakenly detected, so only the bounding box with the highest score is selected. This bounding box is fed to the Segment Anything Model (SAM) \citep{ravi_sam_2024} to extract a mask of the target, as shown in \autoref{fig:segmentation}.

SAM was observed to have fairly accurate results, but had a tendency of leaving holes in the mask or leaving out areas around the object edges. Since the target selection of objects all have mostly convex silhouettes, convex hulls were applied to all masks, because it is important for later steps to have a complete mask of the object, even if parts of the background are included. Of the images where the object was detected, SAM was measured to achieve a Dice score of 0.84. The object mask for each image, $I_i$, is denoted as $M_i$.

\subsection{Segmented 3D Reconstruction}
\label{sec:mtd-reconstr}
In order to recover camera poses, intrinsics, and 3D geometry of the scene, classical photogrammetry techniques are used, which consist of structure-from-motion (SfM) for sparse modeling and multiview stereo (MVS) for dense modeling.

\begin{figure}
    \centering
    \includegraphics[width=.95\linewidth]{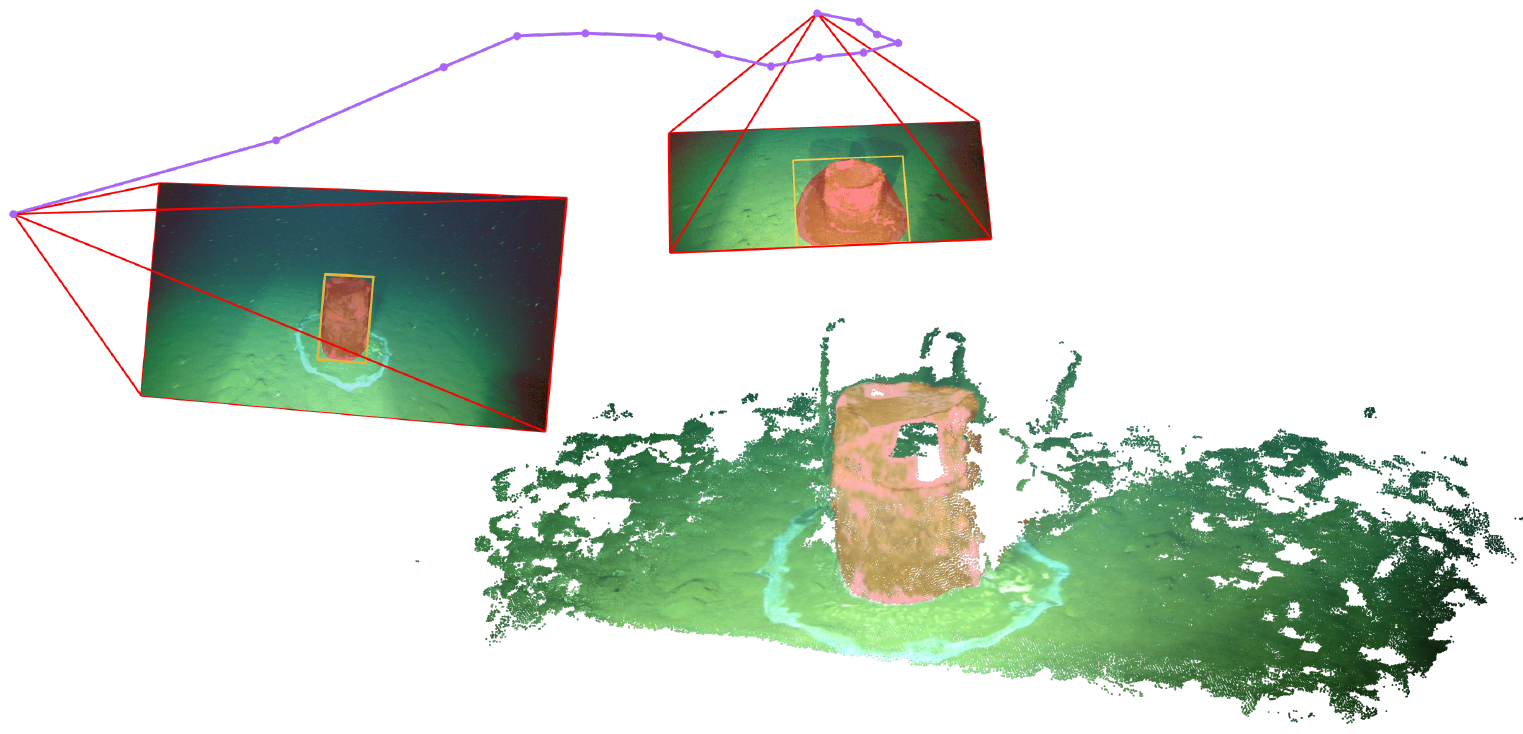}
    \caption{A visualization of the camera trajectory recovered by COLMAP (purple) and the dense 3D reconstruction generated by OpenMVS. Each camera has their corresponding image with a mask and bounding box like in \autoref{fig:segmentation}. The parts of the point cloud segmented out using the SAM segmentation masks are tinted red and represent the object points, as opposed to the rest of the seafloor points.}
    \label{fig:segpc}
\end{figure}

COLMAP \citep{schonberger_structure--motion_2016} is used for SfM, providing the camera intrinsics $\mathbf{K}$, camera poses $\{\mathbf{T}_{\textrm{cam}}^i\}_{i=1}^N$, and sparse point cloud of the scene. Densification is then performed with OpenMVS \citep{cernea_openmvs_2020}, producing a dense point cloud $\mathbf{P}_{\textrm{dense}}$ (\autoref{fig:segpc}).

From the recovered camera poses, intrinsics, and point cloud, every point $p\in \mathbf{P}_{\textrm{dense}}$ is projected into the pixel space for each camera using the pinhole camera model. The number of images where $p$ projects onto the segmented target area in mask $M_i$ is counted. To lower the number of extraneous points segmented for the target, only points that reach a threshold for the number of masks the points project onto are segmented out (\autoref{fig:segpc}). The 3D reconstruction has an inherently ambiguous scale, which must be solved for using the known scale of the target object in later steps of the process (\autoref{fig:scalecorrect}).

This step is also tested with recent learning-based 3D reconstruction methods Fast3r \citep{yang_fast3r_2025} and VGGT \citep{wang_vggt_2025}, which directly predict the camera parameters and dense point cloud as a function of the RGB images.

\subsection{Monocular Coarse Pose Estimation}
\label{sec:mtd-foundpose}
Matching the scaled CAD model to an unscaled point cloud requires an extra DoF from estimating the uniform scale of the point cloud. While methods exist that consider this extra degree \citep{sahillioglu_scale-adaptive_2021}, classical least-squares methods tend to be sensitive to initialization and often get stuck at local minima. FoundPose \citep{ornek_foundpose_2024} is used to obtain coarse monocular object pose estimates in order to use its known dimensions to scale the scene correctly.

FoundPose uses robust visual features from a frozen intermediate DINOv2 layer \citep{oquab_dinov2_2024} to create 2D-3D correspondences between the real object and the synthetic CAD model. DINOv2 is built upon the vision transformer (ViT) \citep{dosovitskiy_image_2021}, meaning it embeds images by splitting them into non-overlapping patches of $14\times14$ pixels and passing them into the network to calculate their patch descriptors. These intermediate layer patches contain a mix of positional and semantic information \citep{amir_deep_2022} that can produce geometrically consistent correspondences \citep{ornek_foundpose_2024}. Each patch is treated as an image feature for the purposes of feature matching.

\begin{figure}[h!]
    \centering
    \includegraphics[width=.8\linewidth]{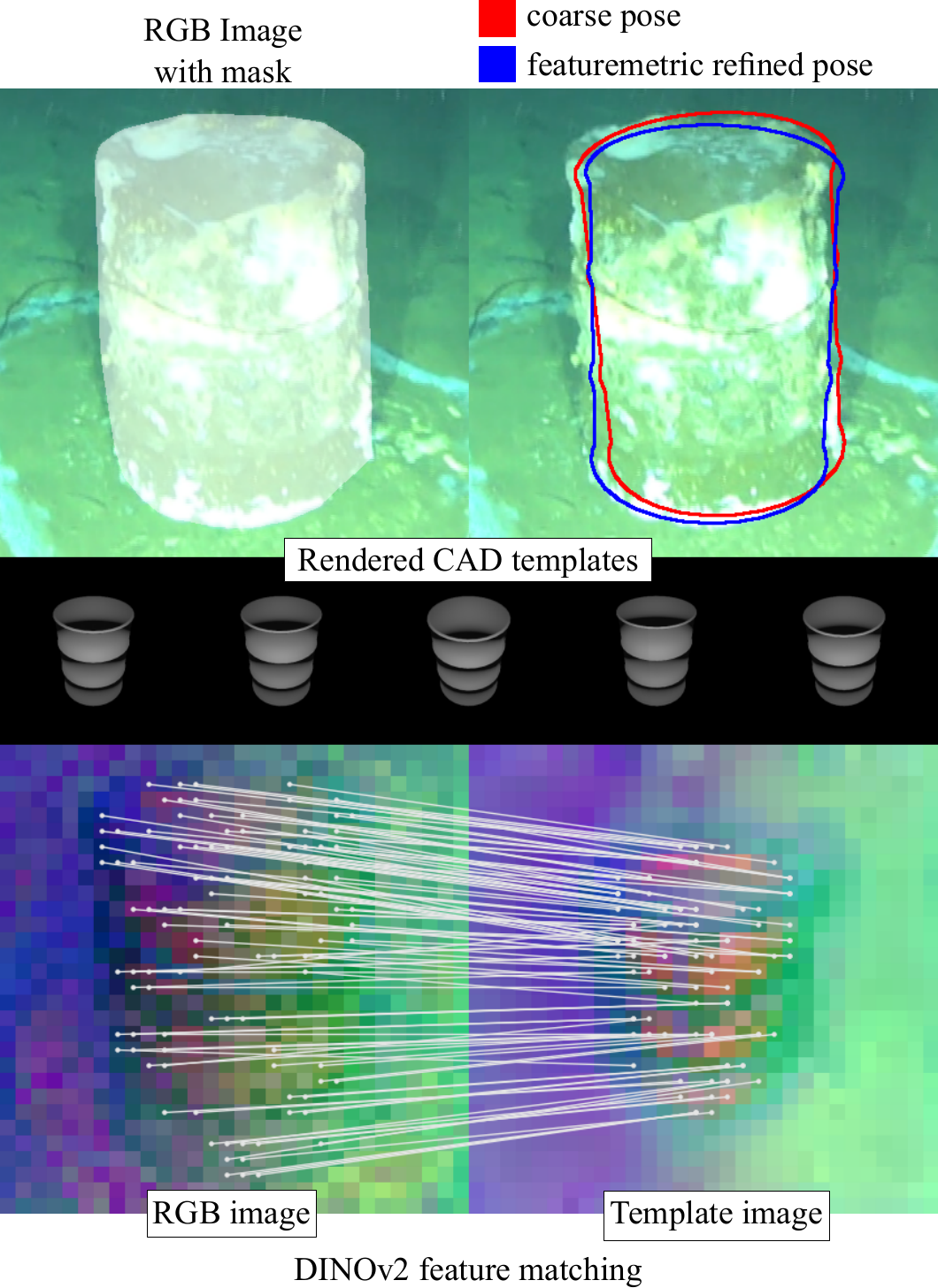}
    \caption{Example FoundPose \citep{ornek_foundpose_2024} intermediary steps and results on a barrel. The input image is shown with its mask overlaid in white (top right), the top five most similar rendered templates (middle row), and the DINOv2 \citep{oquab_dinov2_2024} patch descriptor matches between the real barrel (bottom left) and highest scoring template barrel (bottom right). Each patch descriptor has been projected into RGB space via PCA. Overlaid are the outlines of the predicted coarse pose (red) and featuremetric refined pose (blue) (top right).}
    \label{fig:foundpose}
\end{figure}

As a brief overview of the process, FoundPose renders RGB-D template views of the reference CAD model at uniformly sampled viewing angles and extracts patch descriptors from each. Each template patch descriptor also has a 3D position using the render's depth value in the patch center. The real object in image $I_i$ also has its patch descriptors extracted within the mask $M_i$, and the top $H$ templates with the most similar set of patch descriptors are selected. For each template, the image patches are matched with similar patches in the template, which each have a 3D position, thus creating 2D-3D correspondences. The template object's pose relative to the camera can be obtained by solving the Perspective-$n$-Point (P$n$P) problem. Several algorithms exist to solve P$n$P \citep{lu_fast_2000, lepetit_epnp_2009, terzakis_consistently_2020}, and FoundPose uses efficient P$n$P (EP$n$P) \citep{lepetit_epnp_2009} in conjunction with RANSAC \citep{fischler_random_1981} to obtain a coarse pose. Since each patch is relatively large, further refinement of the pose is done via featuremetric refinement to correct alignment errors, which is described in \citet{ornek_foundpose_2024}. Visualizations of each step are shown in \autoref{fig:foundpose}.

The rest of the pipeline was tested with and without featuremetric alignment, to verify its effectiveness (\autoref{tab:metrics-table}). For the $H$ most similar templates, each hypothesis $j$ provides the monocular pose $_{\textrm{cam}}\mathbf{T}_{\textrm{obj}}^{i,j}$ of the target relative to the camera $i$.

\subsection{Multiview Aggregation of Coarse Poses}
\label{sec:mtd-multiagg}
All buried objects are in various states of degradation, leading to significant differences in texture compared to the reference CAD model. Despite the robustness of DINOv2, FoundPose will suffer degraded performance from these extreme conditions (\autoref{fig:rustyduck}). Since the surface geometry of degraded objects does not differ as much as their texture, it can be used to keep the model robust to texture. Thus, the 3D reconstruction is used to detect and filter geometrically inconsistent poses. The following two sections describe the methodology for aggregating the object poses from FoundPose into the 3D reconstruction world space.
\subsubsection{Resolving Scale Ambiguity}
\begin{figure}
    \centering
    \includegraphics[width=.8\linewidth]{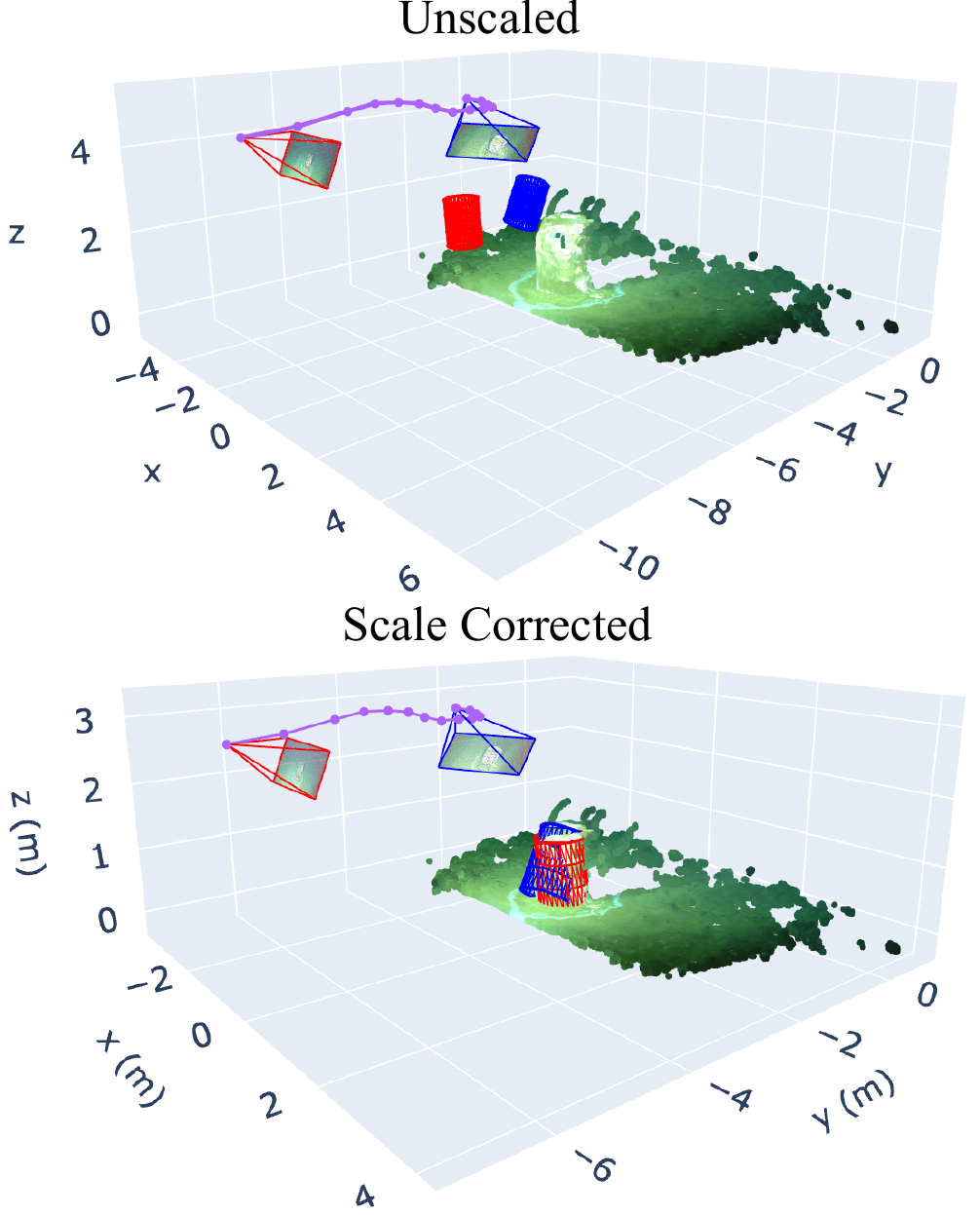}
    \caption{A visualization of the scale correction operation on the photogrammetry 3D reconstruction. Cameras with the trajectory between them in purple are visualized with their own predicted barrel poses as wireframes in their corresponding colors. Unscaled (top) – the scene is too large compared to the camera positions and barrels. Correcting the scale to meters (bottom) keeps the barrel poses relative to the cameras constant, but scales the camera and point cloud positions so that the barrel converges to one position.}
    \label{fig:scalecorrect}
\end{figure}

As shown in \autoref{fig:scalecorrect}, photogrammetric techniques alone cannot resolve the scale of a scene, and require a reference with known size in the scene. Since the object scale is known and monocular pose estimates are available via FoundPose, the scale of the scene and camera translations can be recovered. Scaling a 3D reconstruction only involves scaling the camera translations and point cloud coordinates by a constant, $s$.

Assuming decent estimations for camera poses and object poses in camera frame, a correct scaling factor will cause the centroids of each object to converge near a single point, as shown in the bottom plot in \autoref{fig:scalecorrect}. Thus, solving for this scale can be accomplished by minimizing the variance between the centroids in world frame.

Each object pose $_{\textrm{cam}}\mathbf{T}_{\textrm{obj}}^{i,j}$ corresponding to hypothesis $j$ in camera $i$'s frame can be transformed into the unscaled world frame as follows: $\mathbf{T}_{\textrm{obj}}^{i,j}=(\mathbf{T}_{\textrm{cam}}^{i})(_{\textrm{cam}}\mathbf{T}_{\textrm{obj}}^{i,j})$. The translation components of the poses $\mathbf{T}_{\textrm{obj}}^{i,j},\mathbf{T}_{\textrm{cam}}^i$ are defined as $\mathbf{t}_{\textrm{obj}}^{i,j},\mathbf{t}_{\textrm{cam}}^i\in\R^3$ respectively. The covariance trace of the scaled object positions is minimized with the following formulation:
\begin{equation}
    \min_s \text{tr}(\Var(\{ \mathbf{t}_{\textrm{obj}}^{i,j} + \mathbf{t}_{\textrm{cam}}^i(s-1) \}_{i,j})).
    \label{eq:minvar}
\end{equation}
Solving \autoref{eq:minvar} for $s$ has the following closed form solution:
\begin{equation}
    s = -\frac{\sum_{i=1}^N\sum_{j=1}^H(\mathbf{t}_{\textrm{cam}}^i-\wbar{\mathbf{t}_{\textrm{cam}}})^\top(\mathbf{t}_{\textrm{obj}}^{i,j}-\wbar{\mathbf{t}_{\textrm{obj}}})}{H\sum_i^N\|\mathbf{t}_{\textrm{cam}}^i-\wbar{\mathbf{t}_{\textrm{cam}}}\|^2} + 1,
    \label{eq:minvar-sol}
\end{equation}
where the average translations are $\wbar{\mathbf{t}_{\textrm{cam}}}=\frac{1}{N}\sum_{i=1}^N \mathbf{t}_{\textrm{cam}}^i$ and $\wbar{\mathbf{t}_{\textrm{obj}}}=\frac{1}{NH}\sum_{i=1}^N\sum_{j=1}^H \mathbf{t}_{\textrm{obj}}^{i,j}$.

Outlier object translations are removed via RANSAC during this minimization. Each pose is scaled as $\mathbf{T}=[\mathbf{R}|s\mathbf{t}]$ with original rotation $\mathbf{R}$ and translation $\mathbf{t}$, and the point clouds are scaled by scaling each individual point by $s$. The new scaled camera poses are defined as $\widetilde{\textbf{T}}_{\textrm{cam}}^i$, the scaled object poses as $\widetilde{\mathbf{T}}_{\textrm{obj}}^{i,j}$, the scaled seafloor transform as $\widetilde{\mathbf{T}}_{\textrm{floor}}$, and the scaled corrected point clouds as $\widetilde{\mathbf{P}}_{\textrm{obj}},\widetilde{\mathbf{P}}_{\textrm{floor}}$ for the target object and seafloor respectively.
\subsubsection{Fine Adjustments to Pose}
\begin{figure}
    \centering
    \includegraphics[width=0.8\linewidth]{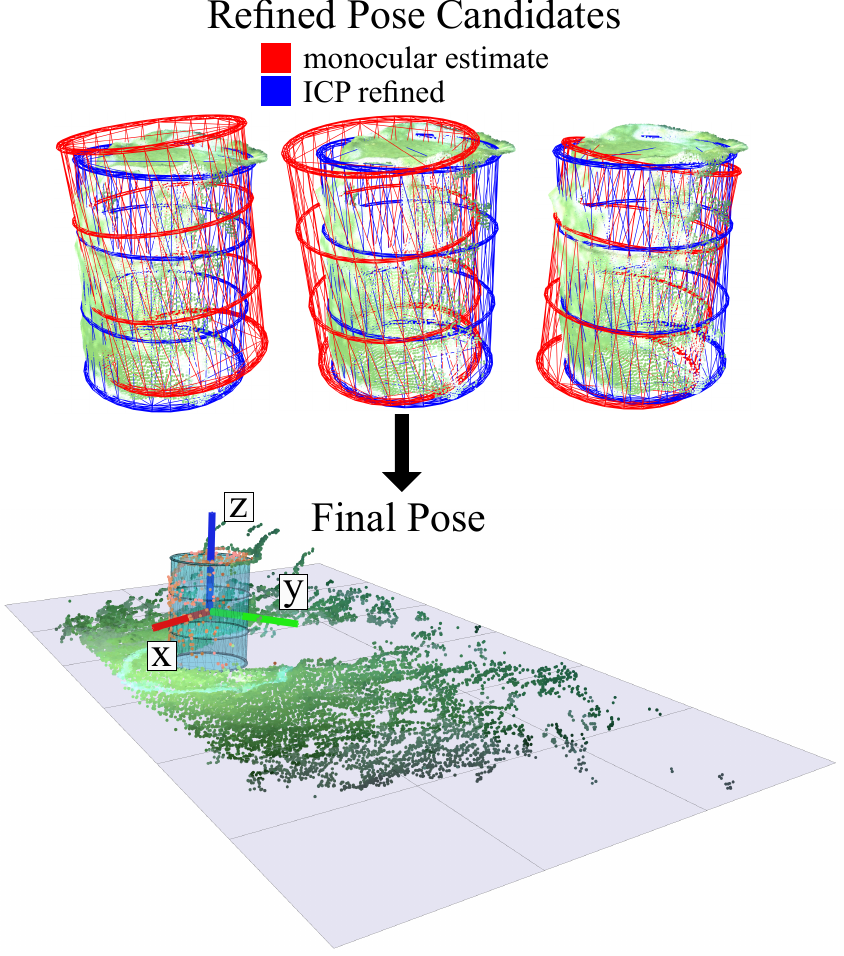}
    \caption{After scale correction, ICP is run on all coarse poses (top), with the original pose (red) and pose after ICP (blue) matched with the barrel reconstruction points (original colors). These poses are symmetry-corrected and averaged to produce a single final pose estimate (bottom), where the barrel mesh (blue) and pose axes are visualized, along with a plane being fit to the seafloor points (purple).}
    \label{fig:fine-angle-ref}
\end{figure}

To further enforce geometric consistency, each pose, $\widetilde{\textbf{T}}_{\textrm{obj}}^{i,j}$, is refined using iterative closest point (ICP) \citep{besl_method_1992}, and a final pose is generated by selecting the best refined poses via RANSAC (\autoref{fig:fine-angle-ref}).

For each camera and hypothesis, points are sampled from the CAD model's surface uniformly and its pose initialized with $\widetilde{\mathbf{T}}_{\textrm{obj}}^{i,j}$. ICP provides a new pose $\mathbf{T}_{\textrm{obj}}^{*i,j}$ from this initialization that most closely aligns the CAD model points with $\widetilde{\mathbf{P}}_{\textrm{obj}}$ (\autoref{fig:fine-angle-ref}), where the FoundPose initialization makes it less likely for ICP to get stuck at a local minimum. Outliers are accounted for in $\widetilde{\mathbf{P}}_{\textrm{obj}}$ by excluding any points that register to a mesh point with a distance greater than $t$ standard deviations from the mean registration distance.

As a final estimate of object pose, the rotation of $\mathbf{T}_{\textrm{obj}}^{*i,j}$ is represented as a unit quaternion $\mathbf{q}_{\textrm{obj}}^{*i,j}\in\mathbb{H}_*$, and the normalized average quaternion of all rotations is taken via the eigendecomposition of the outer product of all quaternions, as described in \citet{markley_averaging_2007}. The average rotation is denoted by $\mathbf{q}_{\textrm{obj}}^*$. The inlier quaternions selected for this average are determined via RANSAC.

To handle object rotational symmetries, $\mathbf{q}_{\textrm{obj}}^{*1,1}$ is selected as a reference rotation, and all other quaternions are rotated to their symmetry-equivalent orientations such that the angle between $\mathbf{q}_{\textrm{obj}}^{*1,1}$ and $\mathbf{q}_{\textrm{obj}}^{*i,j}$ is minimized. For example, if an object has a rotational symmetry about the $z$-axis, then all rotations $\mathbf{q}_{\textrm{obj}}^{*i,j}$ are rotated about the $z$-axis to minimize the angle with the reference rotation. The mean quaternion is then computed in the standard way.

Averaging the translations $\mathbf{t}_{\textrm{obj}}^{*i,j}$ of inlier poses gives average translation $\mathbf{t}_{\textrm{obj}}^*$. Converting $\mathbf{q}_{\textrm{obj}}^*$ to rotation matrix $\mathbf{R}_{\textrm{obj}}^*$, gives the final object pose $\mathbf{T}_{\textrm{obj}}^*=[\mathbf{R}_{\textrm{obj}}^*|\mathbf{t}_{\textrm{obj}}^*]$ (\autoref{fig:fine-angle-ref}), which relative to each camera $i$ is $_{\textrm{cam}}\mathbf{T}_{\textrm{obj}}^{*i}=(\widetilde{\mathbf{T}}_{\textrm{cam}}^i)^{-1}(\mathbf{T}_{\textrm{obj}}^*)$.

\subsection{Plane Fitting and Burial Estimation}
The 3D reconstruction is generated in an arbitrary orientation, but the desired one is such that the scene points up to the positive $z$-axis and the seafloor lies at $z=0$, since this will make burial calculations trivial.

\begin{figure}
    \centering
    \includegraphics[width=0.7\linewidth]{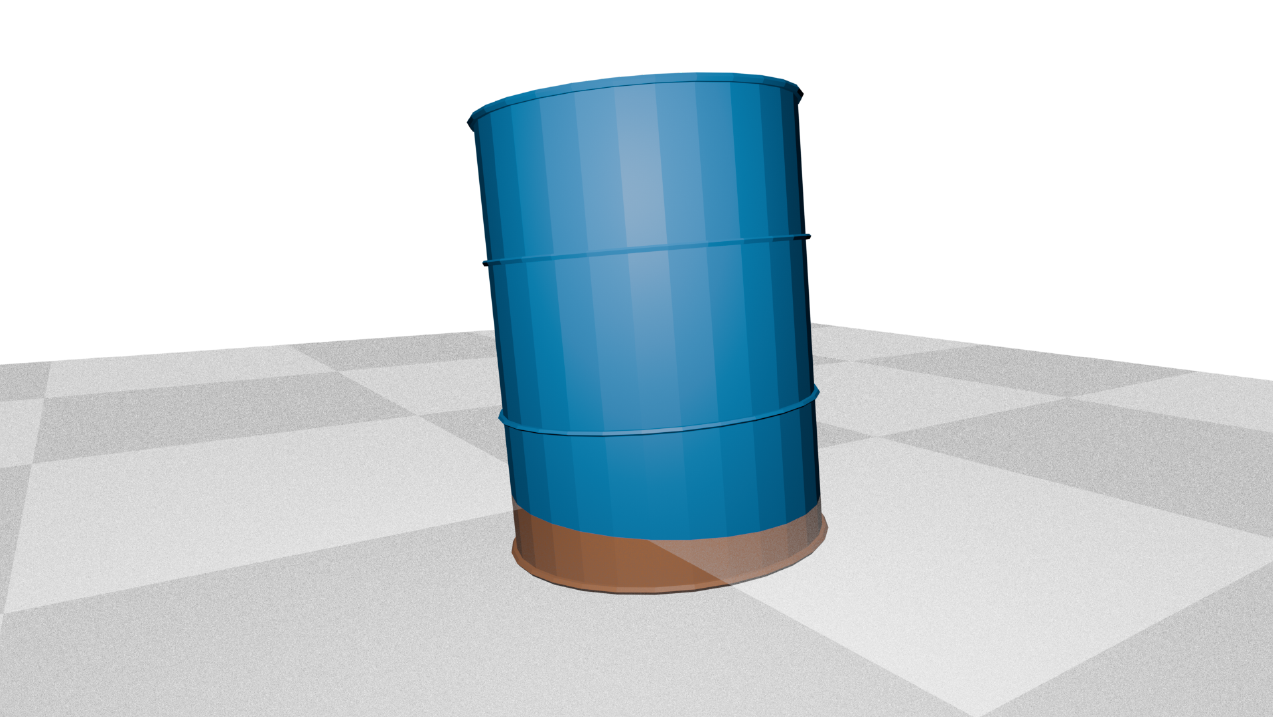}
    \caption{The final render of the predicted barrel pose relative to the fitted plane. The blue part of the barrel mesh is considered above the seafloor, while the orange part of the mesh is considered buried. The lowest point of the buried part is calculated as the burial depth.}
    \label{fig:plane-fit}
\end{figure}

The seafloor is modeled as a plane, fitting $\widetilde{\mathbf{P}}_{\textrm{floor}}$ according to inliers chosen by RANSAC. With this plane, the CAD model and the entire scene can be transformed such that its up direction aligns with $[0,0,1]$ and the floor is located at $z=0$ (\autoref{fig:plane-fit}). A point on the mesh surface, $p_{\textrm{bot}}$, is chosen with the lowest $z$ value ($z_{\textrm{bot}}$). Then the absolute burial depth of the object is $|z_{\textrm{bot}}|$. The units of this depth will depend on the CAD model's units.

\subsection{Implementation Details}
\label{sec:mtd-impldetails}
This section states the parameter settings used in these experiments. Grounding DINO and SAM were run with default parameters. All FoundPose parameters are the same as described in \citet{ornek_foundpose_2024}, which  produced $H=5$ pose hypotheses for each image. The RANSAC inlier thresholds are 0.15 m, 0.05 m, and 0.2 radians, for the scale correction, plane fitting, and rotation averaging respectively. ICP excludes points that are $t=2$ standard deviations away.

\section{Evaluation}
\label{sec:eval}
The error in absolute burial depth can be trivially calculated with the absolute difference between the predicted depth and the manually labeled depth. Similarly, the error relative to the object's size is calculated from the ratio of the burial depths to the height of the object along the $z$-axis (which depends on its pose), taking the absolute difference between predicted and manually labeled values.

The pose of the object is evaluated using the most popular metrics for 6DoF estimates, which are the components of the BOP metric (BOP-M) \citep{hodan_bop_2020}: Visible Surface Discrepancy (VSD), Maximum Symmetry-Aware Surface Distance (MSSD), and Maximum Symmetry-Aware Projection Distance (MSPD). These metrics are symmetry-invariant, meaning they account for object symmetries where multiple orientations produce visually identical poses.

VSD \citep{hodan_bop_2020} treats poses that are indistinguishable in shape as equivalent by measuring the misalignment of only the visible object surface \citep{liu_deep_2024}. It is expressed as follows:
\begin{equation}
    \begin{split}
        & e_{\textrm{VSD}}(\hat D,\bar D,\hat V,\bar V,\tau) = \\
        & avg_{p\in \hat V\cup\bar V}\begin{cases}
            0, & \text{if }p\in\hat V\cap\bar V\wedge|\hat D(p)-\bar D(p)|<\tau\\
            1, & \text{otherwise}.
        \end{cases}
    \end{split}
    \label{eq:vsd}
\end{equation}
$\hat D$ and $\bar D$ are the distance maps created by rendering a model $M$ transformed by the estimated pose $\hat P$ and ground-truth pose $\bar P$ respectively. Each pixel $p$ contains the distance from the camera center to the 3D point $x_p$ on the transformed model that projects onto $p$. The visibility masks $\hat V$ and $\bar V$, which represent where the object is visible in the given poses, are then used to identify which pixels are relevant to compare. $\tau$ determines the tolerance for the misalignment of $z$ coordinates in each pixel. Thus, VSD gives the ratio of pixels that are misaligned up to a tolerance.

MSSD \citep{hodan_bop_2020} measures which vertex in the object has the highest 3D distance from its corresponding vertex from the ground-truth, while considering equivalent symmetries. This can represent how likely a robot can successfully manipulate the object \citep{liu_deep_2024}. It is defined as:
\begin{equation}
    e_{\textrm{MSSD}}(\hat P,\bar P,S_M,V_M) = \min_{S\in S_M}\max_{x\in V_M}\|\hat Px-\bar PSx\|_2.
    \label{eq:mssd}
\end{equation}
$S_M$ are all the global symmetry transformations for model $M$, and $V_M$ are the vertices of the model $M$.

MSPD \citep{hodan_bop_2020} is similar to MSSD, but measures the greatest error in image space. This is useful for applications like augmented reality, where perceivable discrepancies are more noticeable than errors in depth. It is defined as:
\begin{equation}
    \begin{split}
        & e_{\textrm{MSPD}}(\hat P,\bar P,S_M,V_M) = \\
        & \min_{S\in S_M}\max_{x\in V_M}\|\mathrm{proj}(\hat Px)-\mathrm{proj}(\bar PSx)\|_2.
    \end{split}
    \label{eq:mspd}
\end{equation}
$\mathrm{proj}$ is the 2D projection from object vertices into image space.

An estimated pose is considered correct with respect to one of the BOP error functions $e$ if $e<\theta_e$, where $e\in\{e_{\mathrm{VSD}},e_{\mathrm{MSSD}},e_{\mathrm{MSPD}}\}$ and $\theta_e$ is the error threshold of function $e$ \citep{hodan_bop_2020}. The fraction of object instances considered correct is called the recall. The average recall with respect to error function $e$ ($\textrm{AR}_{\mathrm{VSD}}$, $\textrm{AR}_{\mathrm{MSSD}}$, and $\textrm{AR}_{\mathrm{MSPD}}$) is the average of recall rates for multiple threshold settings of $\theta_e$. The threshold settings used are the standard thresholds used in the BOP challenge, as described in Section 2.4 of \citet{hodan_bop_2020}.

The performance of the model is evaluated by recording its metrics with respect to different key parameters (\autoref{tab:metrics-table}).

\subsection{Manually Labeling Poses}
\label{sec:mtd-gt}
As a point of comparison, manually labeled object and seafloor poses are used to evaluate model performance since ground truth in the dataset is lacking. The footage is recorded in an uncontrolled environment where underwater conditions lower video quality, making labeling per object difficult and time-consuming. As a result, only a small subset of all observed objects in the survey can be evaluated. For each video of an object, photogrammetry is used to reconstruct camera positions and manually orient the object and seafloor in the scene such that they are geometrically consistent from multiple viewpoints. This labeling work was done using Blender \citep{blender_foundation_blender_2024}. This process ties the quality of the labeling to photogrammetry performance, but provides better poses than intuiting them from single images. This also reduces the labeling required per object since the pose does not need to be individually labeled per image. It is important to note these challenging conditions cause sizable variance in manual measurements for each object, which is measured by rerunning photogrammetry on different frames of the video and labeling the object again.

\section{Results}
Footage of 54 different objects in the seabed was selected across multiple dives in the San Pedro survey \citep{merrifield_wide-area_2023} and model performance was evaluated in comparison to manually labeled poses according to the previously defined metrics. Although most of the objects selected are drum barrels, the ability to generalize across object categories is demonstrated by testing on several munitions models (\autoref{fig:results-collage}).

\subsection{Model Performance}
\begin{figure*}
    \centering
    \includegraphics[width=0.99\linewidth]{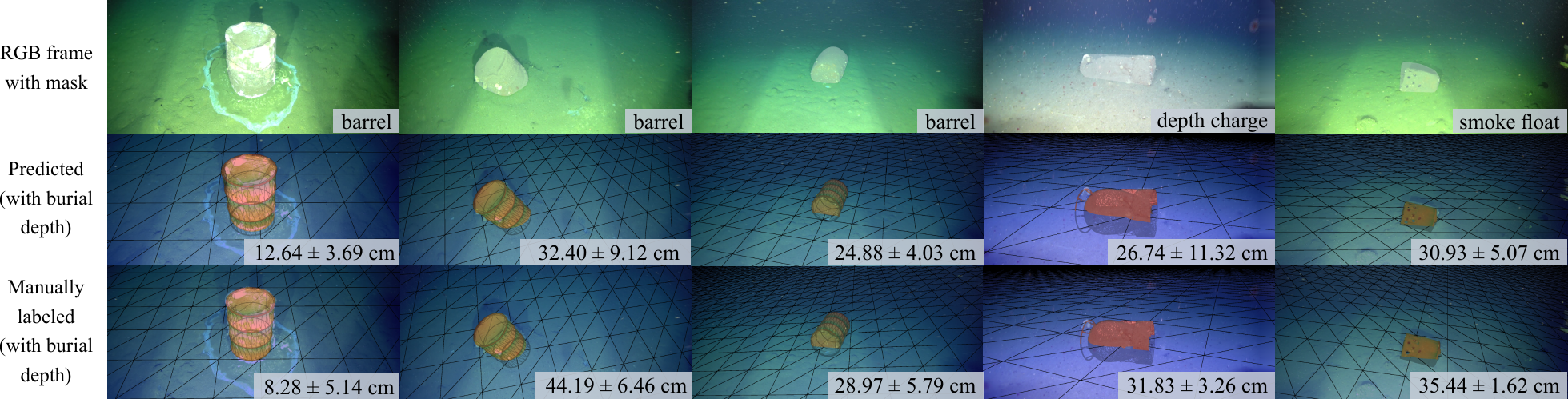}
    \caption{Five example results from footage across multiple ROV dives. Each observed object (column) shows the video frame with the SAM mask in transparent white and object category (top row). The predicted (middle row) and ground truth (bottom row) poses of the object (orange) and seafloor plane (blue) with their wireframe are visualized relative to the ROV camera, along with the calculated object burial depth beneath the plane for each in cm. The uncertainties are the standard deviations of burial depth for $n=6$ separate estimates for predicted and manually labeled depths.}
    \label{fig:results-collage}
\end{figure*}

Illustrated here are several relatively successful examples of pose prediction of objects and seafloor (\autoref{fig:results-collage}). To measure the variance of model-predicted and manually-labeled depths, $n=6$ random samples of images are taken from the video of each selected object, and run through the manual labeling pipeline and predictive model. This allows gauging of the variance in burial depth for both processes. The results are also compared to historically sampled sedimentation rates from a previous study of the San Pedro Basin \citep{alexander_sediment_2009}. Across seven sediment cores sampled from the basin, the vertical accumulation rate of sediment ranges had the following values: $\{0.07, 0.07, 0.10, 0.10, 0.14, 0.14, 0.59\}$ cm yr$^{-1}$. Using these measurements, it is possible to either infer the year an object was dumped using the sampled rates or verify the sedimentation rate itself if the year when the object was dumped is known. Although it is assumed that the object is completely unburied on its initial impact with the seafloor, it is possible to predict initial burial depth using physics-based models \citep{chu_prediction_2005}.

The three barrels in the first three columns of \autoref{fig:results-collage} were found in close proximity to each other in the central highlighted region in \autoref{fig:burial-map}. The upright barrel in the first column has noticeably shallower burial depth, which is likely due to a much shallower initial impact due to it landing on its flat surface. Considering the timespan of dumping took place from 1947--1961, assuming the earliest year means the manually labeled barrel pose yields a burial rate of $0.109\pm0.068$ cm yr$^{-1}$, while the model-predicted barrel pose gives $0.17\pm0.049$ cm yr$^{-1}$, where the uncertainty is standard deviation. These measurements are within the expected sedimentation rate range in the region provided by \citet{alexander_sediment_2009}. The other barrels and objects in \autoref{fig:results-collage} are buried in orientations that would imply a non-negligible initial burial depth, which would overestimate sedimentation rate or object age. Nevertheless, the recovered pose and burial depth of the object can be used to help predict potential mobility of the object.

The model's overall performance is presented in \autoref{tab:metrics-table} according to the previously defined metrics, which demonstrates why enforcing multiview consistency is important for more accurate results. The model's burial depth predictions achieve a mean error of around 10 cm, which is comparatively large given the low yearly sedimentation rates. However, when calculating this error relative to the oriented object height, the error of the ratio of predicted burial depth to oriented object height is 0.11. This means on average, there is only an 11\% error in burial depth relative to the object size. Oriented object height in this case is the length from the lowest to highest point of the object after it has been oriented in the world.

\begin{figure}
    \centering
    \includegraphics[width=.75\linewidth]{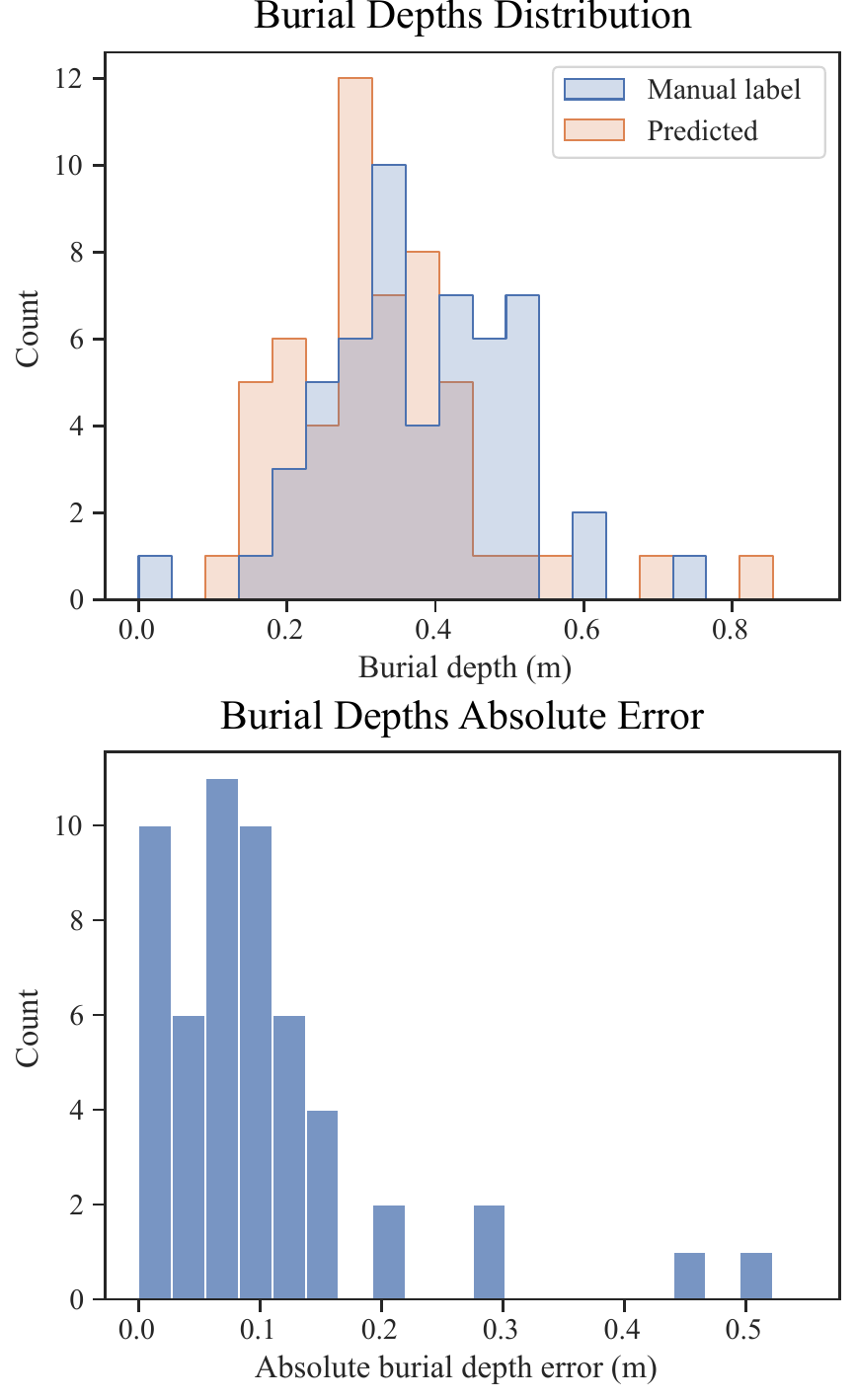}
    \caption{The distributions of ground truth, predictions, and errors for burial depth of $n=54$ different objects.}
    \label{fig:burial-hists}
\end{figure}

\begin{figure}
    \centering
    \includegraphics[width=.75\linewidth]{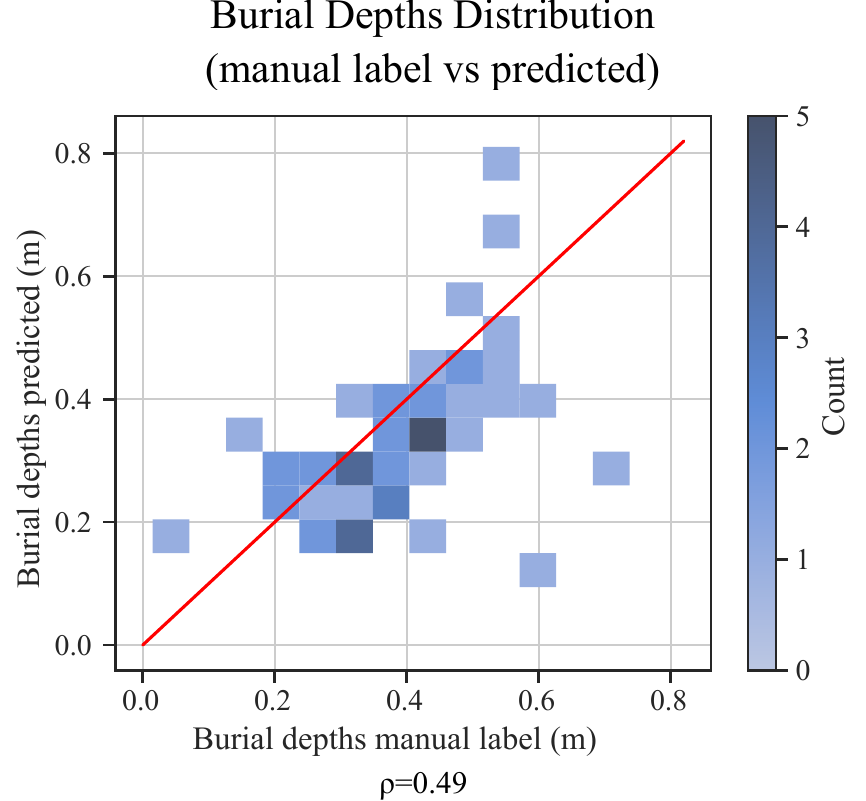}
    \caption{A histogram comparing the manually labeled versus the predicted burial depth of $n=54$ different objects. The red line is $y=x$, where predicted exactly equals ground truth. The trend where the model tends to underpredict depth is visible, and that more extreme errors are linked to when the true burial depth is high; if the object is more buried, there are fewer visible features to use, thus a lower quality pose is generated. The Pearson correlation coefficient for this data is also provided: $\rho=0.49$.}
    \label{fig:burial-hist2d}
\end{figure}

A closer look at the model's error behavior and the distribution of burial depths is provided in \autoref{fig:burial-hists} and \autoref{fig:burial-hist2d}. The majority of objects are between 0.2 and 0.5 m buried, and the model's predictions mostly reflect that behavior. There were only a few catastrophic failures in prediction, and most had under 0.15 m of error (\autoref{fig:burial-hists}). Also observe that the model has a tendency to underestimate the burial depth, and that most high error predictions occur when the object is buried deep (\autoref{fig:burial-hist2d}). The consistent underestimations could be explained by the extra volume every object has from being covered in extra sediment and biological growth, leading to a biased pose prediction since the CAD model would be physically smaller. The higher error at deeper burial depths is expected since more deeply buried objects have fewer features visible, thus leading to more unstable pose estimation (\autoref{fig:fail-case}).

It is important to note that these previously described errors can be difficult to see in \autoref{fig:results-collage}. Slight deviations in the orientation of the seafloor plane are hard to distinguish but can cause sizable errors. Since the seafloor itself has very few distinguishing features, it becomes difficult to localize the seafloor orientation. In addition, closer inspection of the original image can reveal that the contour where the object intersects the seafloor will not always be consistent with the real world. This is because the seafloor is not exactly a plane, and extra sediment tends to build up higher right next to the object. Modeling the seafloor as a plane provides the overall behavior of the sediment in the local region.
\begin{table*}
    \centering
    \caption{Quantitative metrics for model and ablation tests}
    \begin{tabular}{|c|c|c||c|c|c|c|c|}
        \hline
         &  & & & & & Mean burial & Mean burial \\
         & Featuremetric & Reconstruction & & & & depth error & depth ratio \\
        Estimate type & refinement & algorithm & $\textrm{AR}_{\mathrm{VSD}}$ ($\uparrow$) & $\textrm{AR}_{\mathrm{MSSD}}$ ($\uparrow$) & $\textrm{AR}_{\mathrm{MSPD}}$ ($\uparrow$) & (cm) ($\downarrow$) & error ($\downarrow$) \\
        \hline\hline
        FoundPose & no & - & 0.200 & 0.293 & 0.457 & - & - \\
        FoundPose & yes & - & 0.164 & 0.261 & 0.386 & - & - \\
        \hline
        multiview & no & COLMAP & \textbf{0.372} & \textbf{0.471} & \textbf{0.690} & \textbf{10.9} & \textbf{0.110} \\
        multiview & no & Fast3r & 0.128 & 0.182 & 0.278 & 25.6 & 0.198 \\
        multiview & no & VGGT & 0.102 & 0.165 & 0.240 & 25.0 & 0.227 \\
        \hline
    \end{tabular}
    \label{tab:metrics-table}
\end{table*}

\subsection{Mapping Burial Depths}
\begin{figure}[h!]
    \centering
    \includegraphics[width=.9\linewidth]{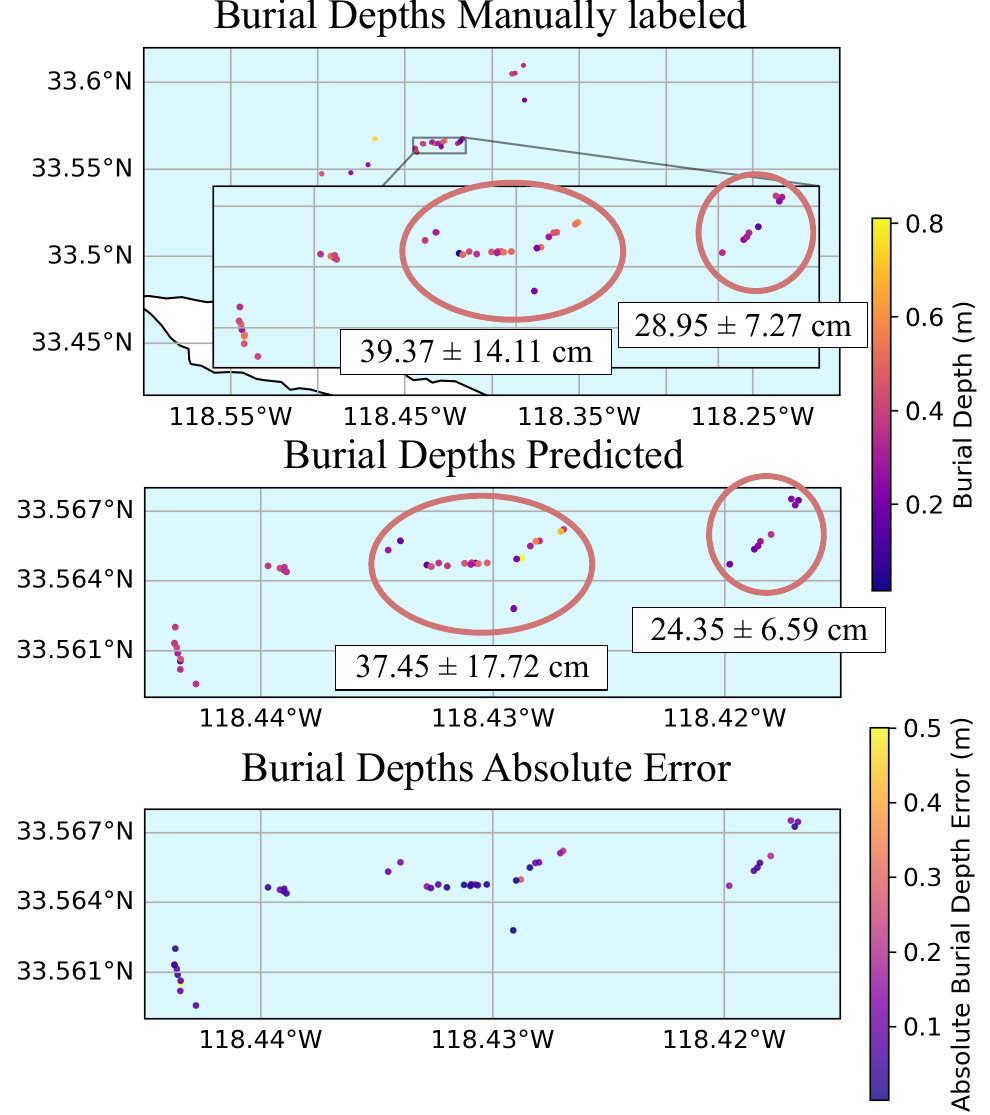}
    \caption{The ground truth, predictions, and errors for burial depth of $n=54$ different objects are shown on the map corresponding to their latitude and longitude. The pipeline configuration with the overall lowest error (\autoref{tab:metrics-table}) is used, and each point is colored according to its burial depth information. Here, the focus is on a smaller, dense area of objects in the survey area approximately 2.8 $\textrm{km}^2$, and regions of interest are circled in red. Notice the objects have a lower burial depth on average over to the east ($n=8$), compared to objects in the center ($n=20$). The depth uncertainty for each region is the standard deviation of objects in that region.}
    \label{fig:burial-map}
\end{figure}

Since latitude and longitude positions are recorded with each object, a map of burial depths can be made across a subset of the search area (\autoref{fig:burial-map}). Specifically, the focus is on a dense area of barrels approximately 2.8 $\textrm{km}^2$. A dense region of barrels is observed in the center of the area ($n=20$ barrels), and it has noticeably higher object burial depth on average (0.39 m) compared to a cluster of barrels to the east ($n=8$ barrels, 0.29 m). The model's predictions reflect this relationship, with a predicted mean burial depth of 0.38 m for the center region, and 0.24 m for the east region. Note that the underestimation bias explained in the previous section still persists.

\subsection{Runtime}
The speed of the model is also measured, excluding the template onboarding stage of FoundPose \citep{ornek_foundpose_2024} (since it only needs to be run once for a given object and camera), and observed to take approximately 280 seconds for 34 images, and 390 seconds for 58 images on a V100-32GB GPU.

\subsection{Ablation Experiments}
Results for ablation experiments are shown in \autoref{tab:metrics-table}. FoundPose's featuremetric refinement \citep{ornek_foundpose_2024} during monocular pose estimation was found to have negligible changes across all metrics, so it is unused for the rest of the experiments.

Different 3D reconstruction methods were also tested in addition to COLMAP \citep{schonberger_structure--motion_2016}, a classical photogrammetry method. Recent deep learning reconstruction models Fast3r \citep{yang_fast3r_2025} and VGGT \citep{wang_vggt_2025} were tested without additional training. As expected, there was a significant decrease in reconstruction quality since these models were mainly trained on terrestrial data. These models were not retrained due to a lack of large-scale, high-quality underwater 3D data.

\section{Discussion}
\subsection{External Burial Factors}
It is important to emphasize that interpreting the burial depth of an object requires consideration of three primary factors: the depth of initial burial, sedimentation rates, and scour from seabed currents. In this study, munitions like depth charges and mousetraps tend to embed themselves deep in the seabed at impact with the seafloor due to their heavy mass and fall velocity, making their burial depths statistically higher on average (\autoref{fig:fail-case}). As a result, nearly all mousetraps and depth charges only had their fins visible, with the majority of the object obscured, which impacted the performance of the model. 

The observed burial states of objects on the seafloor are strongly influenced by the depositional characteristics of the San Pedro Basin, a sedimentary environment shaped by low-energy processes and fine-grained inputs. The basin is predominantly composed of silts and clays with high porosity and variable organic matter content, consistent with long-term accumulation under hemipelagic conditions \citep{berelson_flushing_1991}. These sediment properties not only modulate geochemical gradients but also influence post-depositional alteration and object retention at the seafloor \citep{kemnitz_evidence_2020, zeng_distribution_1996}. Long-term sedimentation rates, determined from $^{210}$Pb and $^{137}$Cs radionuclide profiles as well as varve analysis in low-oxygen settings, have typically ranged from 0.2 to 0.4 cm/yr \citep{kemnitz_evidence_2020, huh_sedimentation_1989, nittrouer_transport_2011}. Such rates are sufficient to partially or fully bury small anthropogenic objects over decadal timescales, depending on local topographic and biological conditions. In particular, the known DDT disposal site in the San Pedro Basin has been subject to this steady accumulation, with rates sufficient to obscure drums and debris dumped during the mid-20th century \citep{kivenson_ocean_2019, us_environmental_protection_agency_region_ix_initial_2021, council_on_environmental_quality_ocean_1970}. Understanding these burial dynamics is essential not only for interpreting acoustic and photogrammetric signatures in modern surveys, but also for reconstructing the historical fate and transport of legacy waste on the seafloor.

While the San Pedro Basin is characterized by fine-grained sedimentation under low-energy depositional conditions, the persistence of some legacy objects resting proud on the seabed suggests that burial is not uniformly achieved. Hydrodynamic scour—often driven by bottom currents or internal wave activity—is likely minimal in this deep basin, as supported by the basin’s laminated sediment structure and relatively undisturbed depositional fabric \citep{berelson_flushing_1991, kemnitz_evidence_2020}. However, localized erosional events, delayed post-depositional processes, or physical interference by object geometry (e.g., drum curvature, corrosion resistance, or low settling velocity) may inhibit complete burial. The presence of World War II-era debris at the sediment surface implies that sediment accumulation alone does not fully dictate burial dynamics; rather, the coupling of biological activity, object properties, and seafloor boundary layer hydrodynamics likely governs long-term exposure. Understanding these interactions remains an important focus for ongoing research.

\subsection{Limitations}
\begin{figure}
    \centering
    \includegraphics[width=0.5\linewidth]{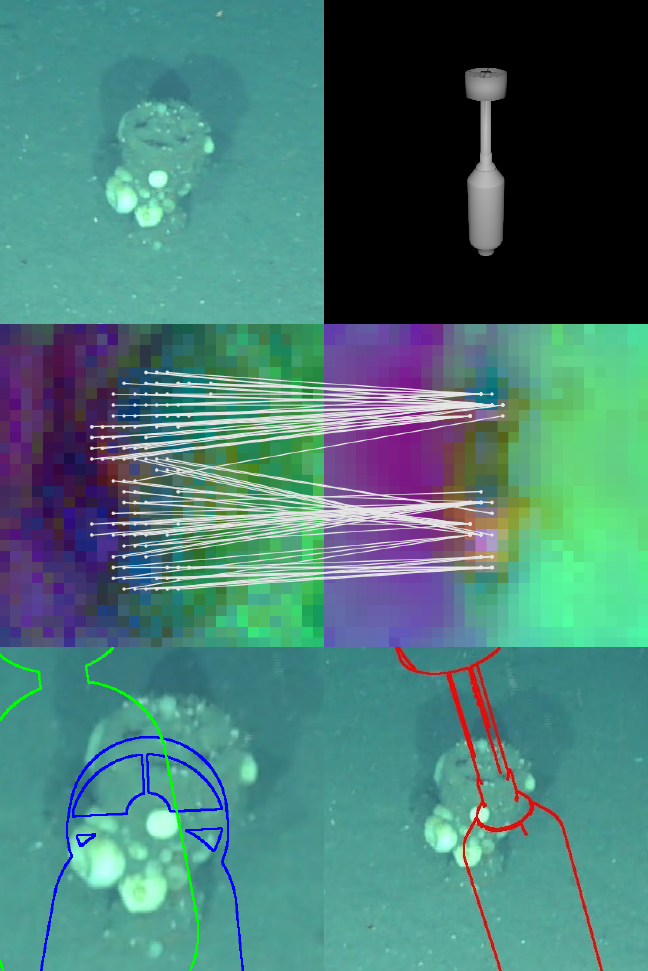}
    \caption{An example failure case with a deeply buried mousetrap, where only the fin is unburied (top left). A decent template was chosen by FoundPose (top right), but the feature matching between the real object (middle left) and template (middle right) had many erroneous matches. Specifically, many features from the fin's biological growth were matched with the head of the mousetrap, which is not visible. This leads to very poor monocular poses (bottom left), both coarse (blue) and featuremetric refined (green), which leads to incorrect multiview refinement (bottom right).}
    \label{fig:fail-case}
\end{figure}

An inherent limitation of PoseIDON is its sequential pipeline structure: errors introduced at any stage propagate forward and compound, since each step depends on the output of the previous one. If early steps produce biased or outlier-corrupted estimates, later steps have no mechanism to recover from them. This is most acutely exposed by the quality of feature matching, which underpins the entire pipeline. Though DINOv2 is robust to texture differences and partial occlusion, reliable correspondence between CAD model renderings and real objects degrades significantly when objects are deeply buried or heavily encrusted with biological growth (\autoref{fig:fail-case}). In such cases, the fraction of incorrect feature matches grows large enough that RANSAC can no longer reliably distinguish inliers from outliers, corrupting the pose estimates that all downstream steps depend on. This failure mode is most prevalent with unexploded ordnance (UXO) such as the mousetrap and depth charge, which tend to fall vertically and embed themselves deep into the seafloor upon impact, leaving few visible and matchable features.

\section{Conclusion}
This study presents a novel computer vision framework, PoseIDON, that enables the estimation of burial depth for anthropogenic seafloor objects using standard ROV imagery. By leveraging foundation model features from DINOv2 and integrating them with photogrammetry and geometric refinement, the method achieves robust 6DoF pose estimation under challenging underwater conditions, including occlusion, object degradation, and variable lighting. The pipeline requires no retraining for new object types, relying instead on CAD models and generalizable visual features. This enables scalable and non-invasive mapping of burial depth across large survey areas.

Evaluated on 54 targets in the San Pedro Basin, including barrels and munitions, the method achieved a mean burial depth error of approximately 10 cm and produced spatial burial patterns consistent with known sedimentation processes. Notably, the model identified systematic underestimation biases and demonstrated that object geometry and initial impact orientation play a critical role in determining observed burial depths.

This approach provides a viable alternative to invasive sediment sampling, enabling large-scale assessments of seafloor burial at legacy waste sites and supporting environmental monitoring with minimal disturbance. The methodology lays the groundwork for future integration with physics-based burial models \citep{chu_prediction_2005} and autonomous survey platforms, offering new opportunities for quantifying sediment dynamics and managing underwater hazards.

\section*{Declaration of Competing Interest}
The authors declare that they have no known competing financial
interests or personal relationships that could have appeared to influence the work reported in this paper.

\section*{Acknowledgment}
This research was supported by the National Oceanic and Atmospheric Administration (NOAA) through a Congressionally directed community project to the Scripps Institution of Oceanography. These funds were allocated to conduct seabed surveys, comprehensively assess contamination from DDT and other pollutants, and explore bioremediation mitigation strategies in the Southern California Bight.
  
We extend our gratitude to the U.S. Navy's Supervisor of Salvage for their instrumental support in executing the fieldwork, particularly through the deployment of the CURV ROV and Trondheim AUV. Their contributions were pivotal in facilitating high-resolution seafloor mapping and the identification of legacy military munitions and industrial waste within the San Pedro Basin.

We also acknowledge the enduring support of the U.S. Office of Naval Research (ONR) under Grant No. N000014-22-C-2006, which has been fundamental in advancing undersea ocean technologies at Scripps. This support has enabled the development and deployment of cutting-edge oceanographic instrumentation, including autonomous underwater vehicles and advanced sensor platforms, which were critical to the success of this study.

The collaborative efforts and sustained support from these agencies have been essential in advancing seabed survey capabilities, enhancing our understanding of the environmental impacts associated with historical ocean dumping practices and in informing potential remediation strategies.

%% The Appendices part is started with the command \appendix;
%% appendix sections are then done as normal sections
\appendix
\section{Additional Analyses of DINOv2 Feature Matching}

\begin{figure}
    \centering
    \includegraphics[width=\linewidth]{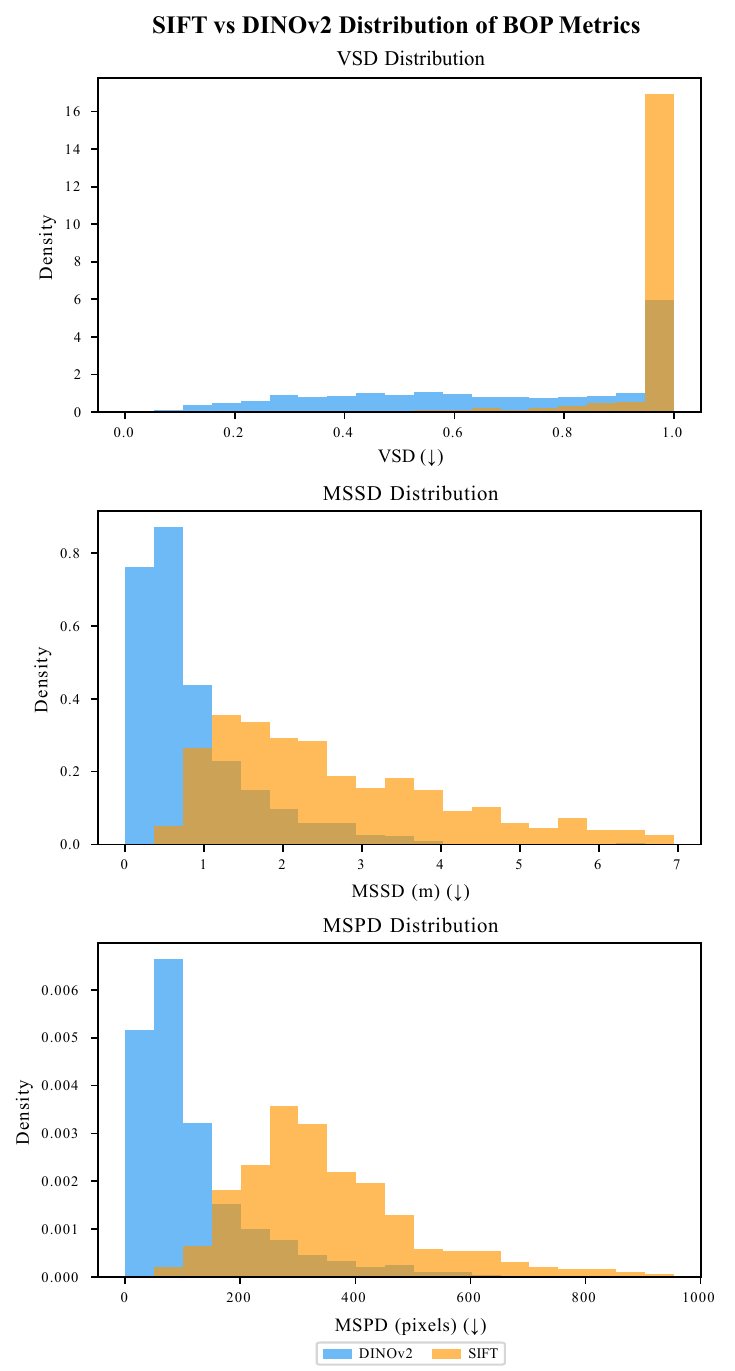}
    \caption{Histograms comparing the distribution of raw BOP metric values (VSD, MSSD, MSPD) for coarse poses estimated using DINOv2 and SIFT features via P$n$P-RANSAC. Each histogram shows results across all video frames matched against their top five templates. DINOv2 produces a concentration of low-error poses, while SIFT yields nearly uniform distributions indicative of effectively random poses, confirming that hand-crafted features cannot bridge the synthetic-to-real domain gap in underwater conditions. Number of DINOv2 poses: $N=1668$. Number of SIFT poses: $N=623$ (the difference in sample size is due to SIFT failing to produce any matches for many frame-template pairs).}
    \label{fig:dino-sift-hist}
\end{figure}
\begin{figure}
    \centering
    \includegraphics[width=\linewidth]{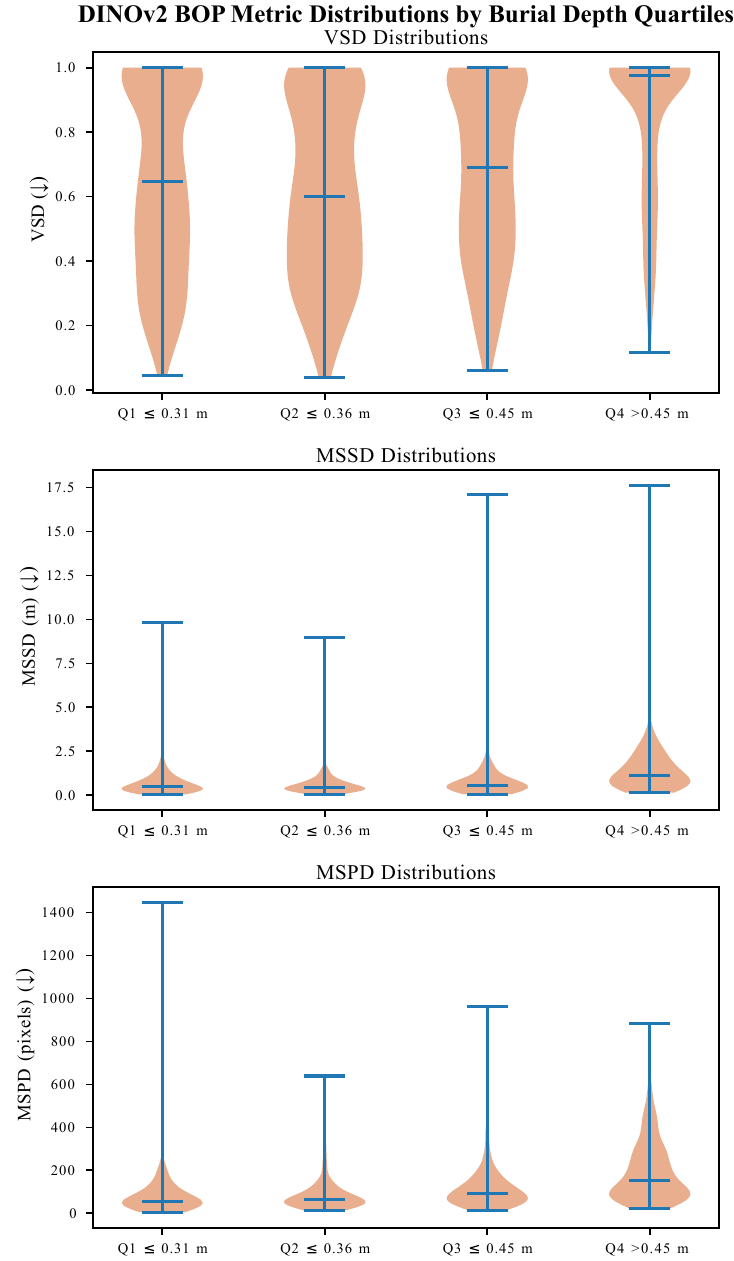}
    \caption{Violin plots of raw BOP metric values for DINOv2-based coarse poses, stratified by burial depth quartile. Bins are separated by 25th, 50th, and 75th percentile. Deeper burial corresponds to greater occlusion and fewer visible features. VSD degrades notably at depths above 0.45~m, where many poses exceed 0.9. MSSD and MSPD show a similar trend: shallow depths produce tightly concentrated low-error distributions, while deeper burial leads to wider spreads toward higher error values. Sample size for each quartile: $N_1=386,N_2=436,N_3=390,N_4=456$.}
    \label{fig:bop-depth-quart}
\end{figure}
\begin{figure}
    \centering
    \includegraphics[width=\linewidth]{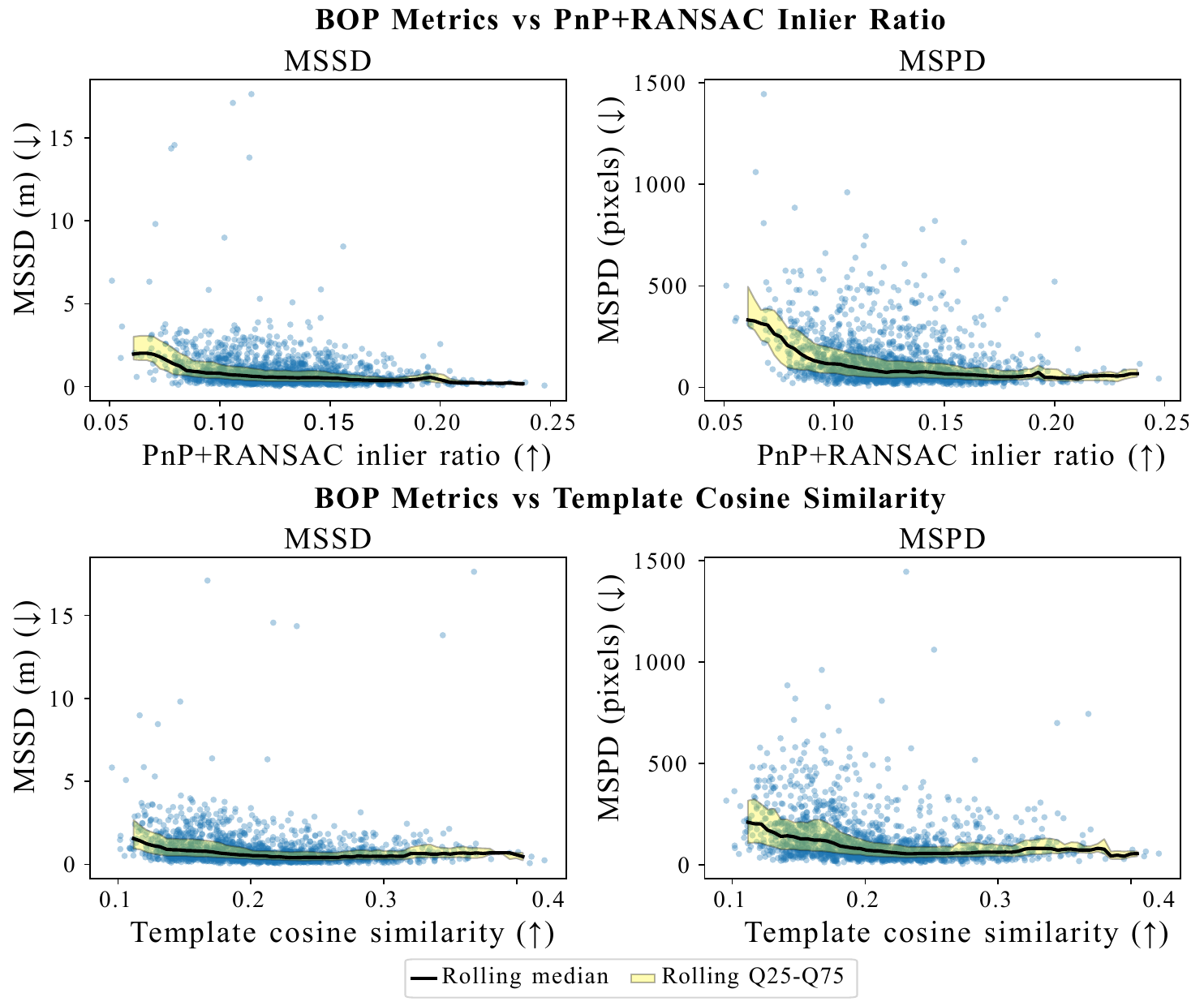}
    \caption{Scatter plots of DINOv2 feature-level statistics versus raw BOP metric values (MSSD and MSPD) for coarse poses. The top row shows P$n$P-RANSAC inlier ratio versus pose error, and the bottom row shows mean cosine similarity between matched DINOv2 patch descriptors versus pose error. In both cases, higher feature quality (more inliers or greater similarity) correlates with lower pose error and reduced variance, indicating that DINOv2 feature reliability directly governs pose accuracy. Trend lines are computed via rolling median and 25th/75th percentile bands. $N=1668$.}
    \label{fig:feature-vs-bop}
\end{figure}

This section provides additional statistical analyses to evaluate the reliability and robustness of DINOv2 as described in Section~\ref{sec:dinov2-robust}. Since ground truth correspondences are not available, feature reliability is measured by comparing the coarse poses produced by P$n$P-RANSAC from FoundPose (Section~\ref{sec:mtd-foundpose}) with manually labeled poses. The BOP metrics described in Section~\ref{sec:eval} are used to evaluate the poses, however the raw values are shown, which provide finer-grained detail on performance than average recall. In this case, lower values indicate better performance. This comparison is done between every video frame and its top five matched templates. Better coarse poses predicted by FoundPose from a single image correlate with more reliable features, since this implies that a sufficient number of features are matched reliably between the underwater domain and rendered template domain.

The performance of SIFT and DINOv2 features during P$n$P-RANSAC is compared as a baseline in~\autoref{fig:dino-sift-hist}. SIFT is unable to handle such a drastic domain gap, which is reflected in the significant difference in distributions with DINOv2, resulting in poor poses. Since DINOv2 can create coherent matches, the majority of DINOv2 poses concentrate at low error values, in contrast to SIFT's wide distribution.

To more specifically evaluate the behavior of DINOv2 features, the coarse pose performance is stratified by burial depth (\autoref{fig:bop-depth-quart}), since it is expected that objects buried deeper are harder to predict due to occlusion. When looking at VSD, the performance is distributed relatively widely between 0 and 1, but at high burial depths ($>0.45 \text{ m}$), many poses are predicted poorly with a VSD above 0.9. MSSD and MSPD also degrade similarly, where the majority of poses are predicted with low error at shallow depths, while at higher burial depths, the distributions spread more widely, with MSSD values reaching well above 2.5~m and MSPD exceeding 200~pixels.

Reliable features should also correlate with a higher inlier count and greater similarity between matched descriptors. As shown in~\autoref{fig:feature-vs-bop}, DINOv2 follows this behavior with respect to MSSD and MSPD. VSD is omitted from this analysis as it incorporates visibility and rendering factors beyond feature matching quality, making it less directly interpretable as a measure of correspondence reliability. When choosing inlier points with RANSAC during P$n$P, a higher inlier ratio indicates more geometrically consistent correspondences, which should correlate with better pose accuracy. This is reflected in the top row of~\autoref{fig:feature-vs-bop}, where the pose error tends to decrease and the spread of values tends to narrow as P$n$P becomes more confident with its matches. 

In addition to the inlier ratio, the cosine similarity between matched DINOv2 patch descriptors in the template and video frame provides a second measure of feature quality, since higher cosine similarity between matched features should also correlate with a higher confidence in the match. The bottom row of~\autoref{fig:feature-vs-bop} shows the same trend, where a higher similarity correlates with better performance and a lower spread of values.

Together, these analyses indicate that DINOv2 features provide reliable cross-domain correspondences for the majority of objects, where degradation primarily occurs at high burial depths due to occlusion limiting the number of visible features. The correlation between feature-level statistics and pose accuracy confirms that DINOv2's cross-domain matching capability is the key factor governing coarse pose quality in this underwater setting.

\section{Analysis of ICP Refinement}

ICP is heavily dependent on initialization and is known to converge to
incorrect local minima under large rotational or translational
errors~\citep{besl_method_1992}. Since PoseIDON uses FoundPose's coarse
estimate as initialization, the quality of ICP refinement is therefore
coupled to the quality of the preceding steps.

To quantify ICP's contribution, we evaluated each dataset with and without
ICP refinement. Table~\ref{tab:icp} summarizes the results.

\begin{table}[h]
\centering
\caption{Effect of ICP refinement across all objects.}
\label{tab:icp}
\begin{tabular}{lcc}
\hline
Metric & Mean change & Std.\ dev. \\
\hline
$\textrm{AR}_{\mathrm{VSD}}$ ($\uparrow$)    & $+0.0087$ & $0.1561$ \\
$\textrm{AR}_{\mathrm{MSSD}}$ ($\uparrow$)   & $+0.0070$ & $0.1434$ \\
$\textrm{AR}_{\mathrm{MSPD}}$ ($\uparrow$)   & $+0.0445$ & $0.2119$ \\
Burial depth error ($\downarrow$) & $-2.6$~cm & $9.94$~cm \\
\hline
\end{tabular}
\end{table}

The results show that ICP's overall contribution is small on average. Mean
improvements are modest across all metrics ($+0.009$, $+0.007$, and $+0.045$
for VSD, MSSD, and MSPD respectively), and the standard deviations are large
relative to the means (0.156, 0.143, and 0.212), indicating that ICP's effect
varies considerably across objects. This suggests that in some cases ICP
refines the pose meaningfully, while in others it has little effect or causes
degradation. Burial depth error shows a similar pattern, with a mean
improvement of 2.6~cm but a standard deviation of 9.94~cm. Taken together,
these results suggest that ICP provides only incremental benefit on average,
which is consistent with its known sensitivity to initialization quality.

% Appendix text.

% %% For citations use: 
% %%       \citet{<label>} ==> Lamport (1994)
% %%       \citep{<label>} ==> (Lamport, 1994)
% %%
% Example citation, See \citet{lamport94}.

%% If you have bib database file and want bibtex to generate the
%% bibitems, please use
%%
\bibliographystyle{elsarticle-harv} 
\bibliography{references}

%% else use the following coding to input the bibitems directly in the
%% TeX file.

%% Refer following link for more details about bibliography and citations.
%% https://en.wikibooks.org/wiki/LaTeX/Bibliography_Management

% \begin{thebibliography}{00}

% %% For authoryear reference style
% %% \bibitem[Author(year)]{label}
% %% Text of bibliographic item

% \bibitem[Lamport(1994)]{lamport94}
%   Leslie Lamport,
%   \textit{\LaTeX: a document preparation system},
%   Addison Wesley, Massachusetts,
%   2nd edition,
%   1994.

% \end{thebibliography}
\end{document}